\documentclass{article}
\usepackage[utf8]{inputenc}

\usepackage{xcolor}
\usepackage{hyperref}
\usepackage{booktabs}
\usepackage{longtable}
\usepackage{amsfonts}
\usepackage{stfloats}
\usepackage{graphicx}
\usepackage[numbers, sort]{natbib}
\usepackage{authblk}
\usepackage{amsmath}
\usepackage{amssymb}
\usepackage{amsthm}
\usepackage[english]{babel}
\usepackage{charter}
\usepackage{fullpage}
\usepackage{pifont}
\usepackage{color,soul}
\usepackage{listings}
\usepackage{lipsum}
\usepackage{empheq}
\usepackage{subfigure}
\usepackage[most]{tcolorbox}
\usepackage{multirow}
\usepackage{tabularx}
\usepackage{makecell}
\usepackage{xspace} 
\usepackage{wrapfig}

\title{On Efficient Training of Large-Scale Deep Learning Models:\\ A Literature Review}

\author{
   Li Shen$^{\S}$\thanks{The first three authors (alphabetical order) contribute equally .}~,\ 
  Yan Sun$^{\dagger*}$,\ 
  Zhiyuan Yu$^{\ddagger*}$,\ 
  Liang Ding$^\S$,\  
  Xinmei Tian$^\ddagger$,\ 
  Dacheng Tao$^\dagger$ \\
     \vspace{0.5em}
$^\S$JD Explore Academy, China\\  
$^\dagger$The University of Sydney, Australia \\
$^\ddagger$University of Science and Technology of China, China\\
   \vspace{0.5em}
   \texttt{mathshenli@gmail.com, ysun9899@uni.sydney.edu.au, yuzhiyuan@mail.ustc.edu.cn}\\
    \texttt{liangding.liam@gmail.com, xinmei@ustc.edu.cn, 
    dacheng.tao@gmail.com}
}

\date{}
\begin{document}

\maketitle

\begin{abstract}
The field of deep learning has witnessed significant progress in recent times, particularly in areas such as computer vision (CV), natural language processing (NLP), and speech. The use of large-scale models trained on vast amounts of data holds immense promise for practical applications, enhancing industrial productivity and facilitating social development.  However, it extremely suffers from the unstable training process and stringent requirements of computational resources. With the increasing demands on the adaption of computational capacity, though numerous studies have explored the efficient training field to a certain extent, a comprehensive summarization/guideline on those general acceleration techniques of training large-scale deep learning models is still much anticipated. In this survey, we present a detailed review of the general techniques for training acceleration. We consider the fundamental update formulation and split its basic components into five main perspectives:
(1) ``data-centric": including dataset regularization, data sampling, and data-centric curriculum learning techniques, which can significantly reduce the computational complexity of the data samples;
(2) ``model-centric", including acceleration of basic modules, compression training, model initialization and model-centric curriculum learning techniques, which focus on accelerating the training via reducing the calculations on parameters and providing better initialization;
(3) ``optimization-centric", including the selection of learning rate, the employment of large batchsize, the designs of efficient objectives, and model average techniques, which pay attention to the training policy and improving the generality for the large-scale models;
(4) ``budgeted training", including some distinctive acceleration methods on source-constrained situations, e.g. for limitation on the total iterations;
(5) ``system-centric", including some efficient distributed frameworks and open-source libraries which provide adequate hardware support for the implementation of above mentioned acceleration algorithms.
By presenting this comprehensive taxonomy, our survey presents a comprehensive review to understand the general mechanisms within each component and their joint interaction. Meanwhile, we further provide a detailed analysis and discussion of future works on the development of general acceleration techniques, which could inspire us to re-think and design novel efficient paradigms.
Overall, we hope that this survey will serve as a valuable guideline for general efficient training.
\end{abstract}

\section{Introduction}
  
\begin{figure}[t]
\centering
\begin{minipage}[t]{0.5\textwidth}
\centering
\includegraphics[width=0.99\textwidth]{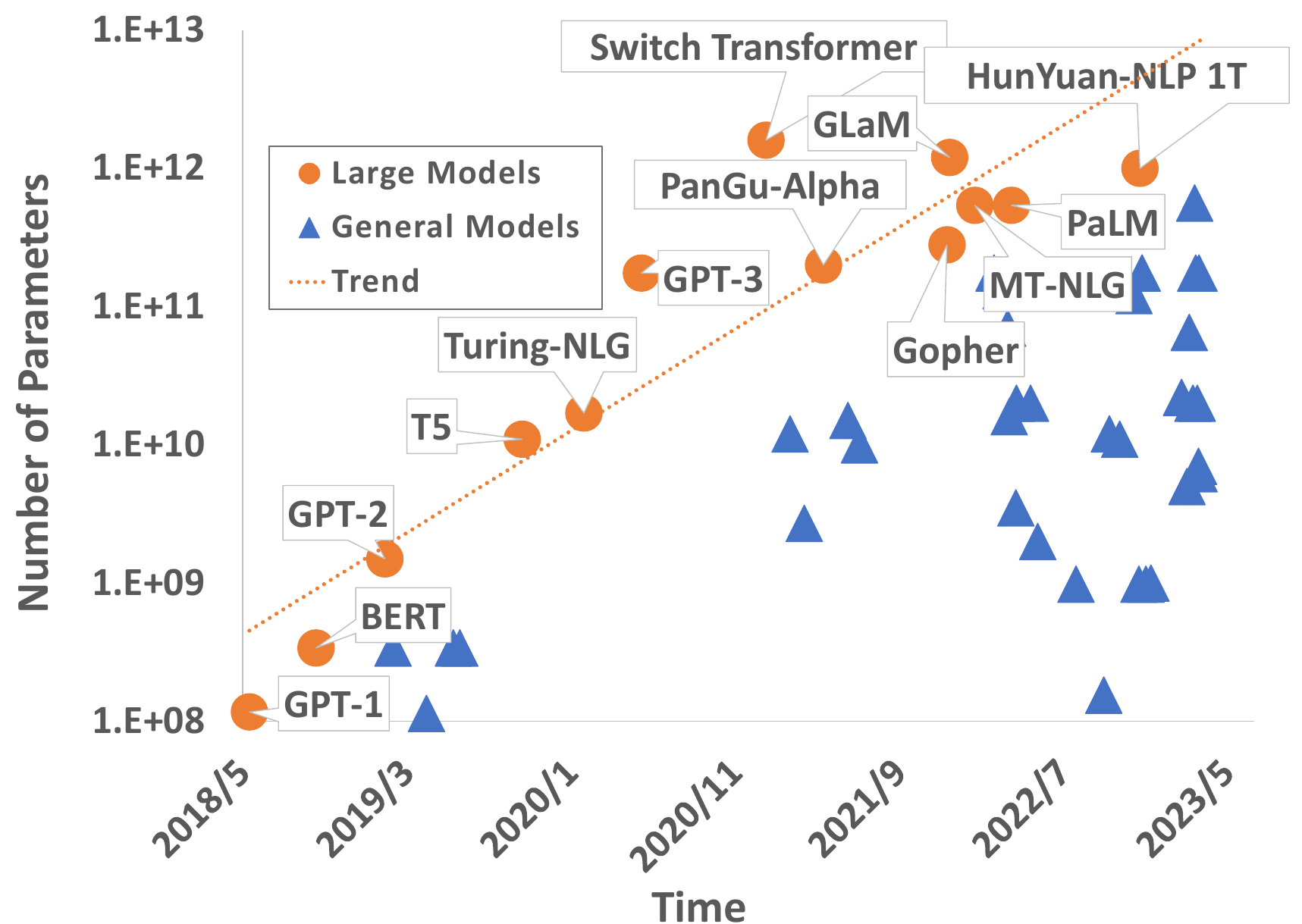}
\label{fig:milestone}
\vspace{-0.2cm}
\caption{Milestone of large-scale models.}
\end{minipage}\!\!
\begin{minipage}[t]{0.5\textwidth}
\centering
\includegraphics[width=0.99\textwidth]{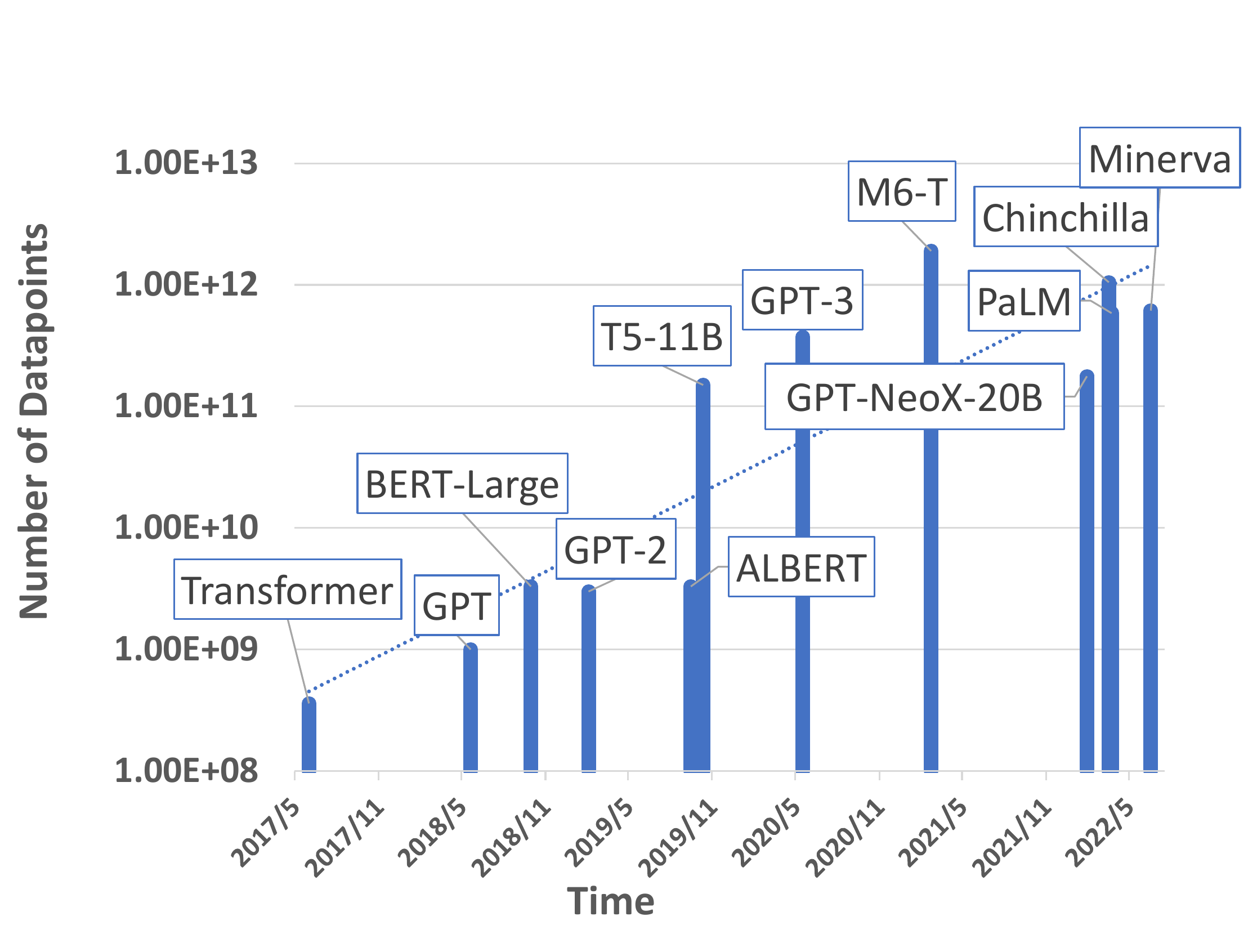}
\label{fig:pretrain}
\vspace{-0.2cm}
\caption{Milestone of the large datasets.}
\end{minipage}
\end{figure}

With the rapid development of artificial intelligence techniques, parameters of deep models have a remarkable growth spurt with millions and even billions. \citet{kaplan2020scaling} study the relationships of model size, dataset size, and the amount of computation used for training as a power-law and indicate that larger models are inherently data-hungry and significantly more sample-efficient on the learning. The deployment of large models has also become one of the most important research fields. For instance, \citet{dehghani2023scaling} propose the ViT-22B which demonstrates the potential for ``LLM (large language model)-like" scaling in the computer vision~(CV) community. GPT-1~\citep{radford2018improving} proposes the supervised fine-tuning to drive the language model with $0.1$B parameters. While two years later, GPT-3~\citep{brown2020language} holds $175$B parameters trained on $45$TB data samples and successfully achieves state-of-the-art results on various natural language processing tasks. Turing-NLG employs the generative language model with approximately $17.2$B parameters and it takes only one year to quickly iterate to a sizeable model MT-NLG with $530$B parameters~\cite{smith2022using}, which is well ahead of the GPT-3 in several tasks. We summarize the milestone of the development of model sizes proposed with time in Figure~\ref{fig:milestone}. Even though the gains from this rapid growth are stunning, to sustain practical efficiency, substantial progress in exploring novel techniques and training capabilities is eagerly awaited. For now, the enormous and expensive costs of training such sizeable models are usually unacceptable. Specifically, training the GPT-3 consumes approximately $355$ GPU years at a cost of $\$4.6$M. Under such a huge amount of parameters and data samples, traditional training from scratch clearly can not afford the massive expense, especially when expanding to the downstream tasks~\citep{pang2023simpletrack,li2022mhformer,cheng2022masked,rogers2023qa,paramasivam2022survey,tan2022adversarial}, which introduce additional architectures and excessive parameters. Therefore, the pretraining-finetuning mode has increasingly drawn lots of attention and shines in the field of deep learning.

Inheriting the paradigm of transfer learning, the leitmotif of pretraining is training a general foundation model on a common dataset to obtain an outstanding feature extractor~(also a good initial state from the perspective of optimization), to help to implement and stabilize the training process on another specific task. Many research areas benefit from the pretraining approach. Concretely, it contributes to training the large transformer models in the CV community by applying the recovered randomly masked images. Masked autoencoder~(MAE)~\citep{he2022masked} develops an asymmetric encoder-decoder architecture to couple the self-supervised reconstruction and backend training, yielding a promising transfer performance for the downstream tasks. Likewise, bidirectional encoder representation from image transformers~(BEiT)~\citep{bao2021beit} follows the BERT~\citep{devlin2018bert} to reconstruct the images via the visual tokens generated from the blockwise masked images. In the NLP community, \citet{radford2018improving} propose the generative pretraining~(GPT) to realize large gains by adopting the generative pretraining on the diverse corpus of unlabeled text. To further improve its efficiency, GPT-2~\citep{radford2019language} greatly increases the model capacity as an extractor of word vectors with extensive semantic knowledge. GPT-3~\citep{brown2020language} studies in-context learning via a double-loop strategy which could significantly enhance the understanding of knowledge in the pretraining and promote fluidity and generality in the practical scenes. \citet{hendrycks2019using} study that using pretraining can improve the model robustness and uncertainty, which shows great advantages for training on a huge dataset to generalize well. Recent advances reveal the substantial gain from the pretraining models with huge capacity, especially on task-agnostic and few-shot scenarios. It also presents a promising direction for future development, which is, the large-scale model with enormous parameters pre-trained on the extensive dataset is able to exceed the performance of other supervised training. This exciting progress efficiently reduces the dependence on the ground truth in deep learning and greatly inspires the application of big data. Meanwhile, it also puts more stringent requirements on computation and training efficiency. Expensive costs including both time and money seriously hamper its development. In this context, we study this comprehensive review to introduce and summarize the general and practical techniques for the training acceleration on the large-scale models, which have promoted faster training and may also contribute to the huge foundation model in the pretraining.

Many recent works review and summarize the efficient training techniques, mainly including the introduction of efficient pre-training models, the novel designed components for acceleration, advanced optimization methods, efficient training on the NLP or CV communities, and the bag of tricks in the training process. \citet{qiu2020pre} present a review and a systematically categorized list of pre-trained models for solving the various NLP tasks. They study the development history of language models and current advances in the pre-training.
\citet{han2021pre} summarize the research works related to the general pre-training techniques and provide some insights into their future research.  
\citet{bommasani2021opportunities} introduce the efficient foundation models, mainly from the perspectives of their general concepts, their powerful capabilities, their underlying training techniques, and applications. They also summarize the evolution of the pre-training and the current challenges in the practical scenarios. 
\citet{zhou2023comprehensive} provide a comprehensive review of pre-trained foundation models (PFM), discussing the recent research advances in their achievements on each different community which may bring novel thinking to the local studies. Meanwhile, they systematically summarize some of the main problems and future challenges in the application.
\citet{lin2022survey} focus on the novel Transformer models and review the several variants of the Transformer models, which are aligned to consider the efficient architecture modification, pre-training techniques, and training accelerations. 
\citet{weng2023transformer} review the development of the Transformer family, which introduces the detailed evolution of the Transformer models and systematically analyzes the strengths and weaknesses of each architecture.
\citet{tay2022efficient} summarize some variants of the efficient Transformer models. From the view of practical training, they provide some strategies to improve the efficiency of training Transformer models and suggestions for future studies. 
\citet{zhuang2023survey} study the overview of the efficient training of Transformers including computation efficiency, memory efficiency, and hardware/algorithm co-design. Different from them, we focus more on the fundamental acceleration techniques which are not limited to the Transformer models.
\citet{openmlsys2022} provide a comprehensive study on the designing and implementation of efficient machine learning systems. They pay more attention to the practice of data pre-processing, forward and backward computation, efficient deployment and communication in parallel, and specific implementation of the optimization methods.
\citet{he2021large} study the recent advances of large-scale deep learning in terms of generalization guarantee and optimization efficiency, which include the novel optimizers and strategies to address the training overheads and reduce the required memory in the computational devices. They also elaborate on the exploration of large-batch training. 
\citet{he2019bag} summarize a bag of tricks for training CNN models. They conduct systematic experiments and summarize some effective techniques for data augmentation and the design of the ingenious learning rate scheduler. 
\citet{treviso2022efficient} summarize efficient methods for NLP and discuss their efficiency and shortcomings.

Recently, efficient training of large-scale deep learning models has become a critical research area in machine learning. While there has been significant progress in this field, much of the existing studies focus on specific model architectures or serve particular communities. In contrast, our study offers a comprehensive review of practical acceleration techniques for any large-scale deep learning model, independent of the task or model architecture. From the perspective of practical efficiency, we consider that efficient training mainly focuses on two clear targets:
\begin{itemize}
    \item \emph{To achieve comparable test accuracy, efficient training requires less training time.}
    \item \emph{Under similar training costs, efficient training achieves higher performance.}
\end{itemize}

Our survey provides insightful guidance on the general training acceleration of deep learning models. We analyze the efficacy of training acceleration techniques on various basic backbone architectures that underpin many modern deep learning models. By examining different architectures of deep networks, our review can help achieve efficient training across any type of deep learning model. Additionally, since our survey is task-free and model-free, it provides a broad generalization of training acceleration techniques that can be applied across different domains and model architectures. Our survey aims to be a helpful resource for researchers and practitioners looking to accelerate the training of their large-scale deep learning models. By understanding the general principles behind effective training acceleration techniques, researchers can develop faster and more efficient models without being constrained by specific architectures or tasks. Overall, our study makes a significant contribution to the field of machine learning by providing a comprehensive survey of general training acceleration techniques for large-scale deep learning models.

In our survey, we are interested in solving the general fundamental minimization problem which could be easily expanded to the training foundation models or pretraining tasks:
\begin{equation}
\label{eq:min_problem}
    \min_{w}\Big\{F(w)\triangleq \mathbb{E}_{\xi\sim\mathcal{D}}\big[ f(w;\xi)\big]\Big\},
\end{equation}
where $w$ is the parameters, and $\xi$ represents the data samples subject to $\mathcal{D}$. On the practical scenarios, we consider a large training set $(X,Y)$ and utilize the empirical risk minimization~(ERM) loss to approximate the full expectation $F$. In this case, each single data pair $(X_i,Y_i)$ is sampled uniformly from the training set. Thus we solve the following problem:
\begin{equation}
\label{eq:ERM}
    \min_{w}\Big\{\bar{F}(w)\triangleq \frac{1}{m}\sum_{i=1}^m f_i(w)\Big\},
\end{equation}
where $f_{i}(w)\triangleq f(w;(X_i,Y_i))$ represents the loss of the $j$-th data sample, and $m$ the size of the dataset $(X,Y)$. When $m$ is large enough, the ERM approaches the full expectation, e.g. for $m\rightarrow \infty$, $\bar{F}\rightarrow F$.\\



Different from the previous works, we deconstruct the general \textit{gradient-based} descent formulation as the architecture of this review. Specifically, we consider all the components in the fomulation~(\ref{eq:gradient-based descent}) which could cover the total training process in the deep learning. We omit the additional proximal terms by absorbing them into $f$. Without loss of the generality, we use the update vector $G$ instead of the gradients to encompass a wide range of methods. We consider the fundamental update formulation as:
\begin{equation}
\label{eq:gradient-based descent}
    w_T=w_0-\sum_{t=0}^{T-1}\frac{\gamma_t}{B}\sum_{i=1}^{B}{G(w_t;(X_i,Y_i))}.
\end{equation}
In Table~\ref{tab:directory}, we summarize the notations and their corresponding research field. Based on Equation~(\ref{eq:gradient-based descent}), by refining and splitting the different roles of components, we distinguish the previous works according to their inherent heuristic insights and theoretical scenes into $5$ major categories. Each corresponds to the optimized target of the computation efficiency under the classified group. We fine-grain the above components to categorize current general acceleration techniques for training large-scale models with the feasibility of practical implementation. Specifically, they are:

\begin{table}[t]
\centering
\label{tab:directory}
\caption{Notations of the terms in Equation~(\ref{eq:gradient-based descent}). The textual structure of this survey is followed by this table. We disassemble the fundamental update formulation in the training to introduce how the current acceleration techniques are designed and implemented on each term.}
\vskip 0.15cm
\begin{tabular}{clcc}
\toprule
Components & Definition & Research Field & Text\\
\midrule
$(X_i,Y_i)$ & dataset samples & Data-centric & Section~\ref{sec:data}\\
\midrule
\makecell[c]{$w$\\$w_0$} & \makecell[l]{model parameters\\weight initialization} & Model-centric & \makecell[c]{Section~\ref{sec:model}}\\
\midrule
\makecell[c]{$B$\\$\gamma$\\$G$} & \makecell[l]{size of the minibatch\\learning rate\\gradient estimator} & Optimization-centric & \makecell[c]{Section~\ref{sec:optimization}} \\
\midrule
$T$ & maximum iterations & Budgeted Training & Section~\ref{sec:budgeted_training}\\
\midrule
$\sum$ & summation & System-centric & Section~\ref{sec:system}\\
\bottomrule
\end{tabular}
\end{table}

\begin{itemize}
    \item \textbf{Data-centric efficient training.}
    In deep learning, there is often a gap between global expectations and the distribution of training samples. This can lead to improvement in test accuracy during the middle and late stages of training, despite efficient performance in the early stages. To address this issue and improve generalization performance, data-centric methods are employed to expand the sample capacity of the training set via effective data augmentation and regularization policies. It requires additional pre-processing calculations to enhance the diversity and maintains a potential for higher stability, leading to better generalization performance in real-world applications. Meanwhile, to achieve the efficient accelerations and further improve the generality of models, data-centric methods study the valid sampling techniques to select a key sub-set during stochastic optimization. It effectively reduces the number of samples that are required to be involved in calculating the gradients. Furthermore, it protects the models from over-fitting in training at those unimportant samples or data that have been learned well enough. Recent advances learn that curriculum learning provides a progressive process, yielding a efficient training. It uses the low-resolution samples with less regularization at the early stage of training and gradually recover them to the high-quality samples. In summary, the core consideration of data-centric methods is how to reduce data processing requirements without compromising the performance.

    \item \textbf{Model-centric efficient training.} 
     The deep model is an elaborate mapping function from the data domain to the ground truth. Past works have explored a number of mature architectures to construct a network for efficient training, e.g. convolution-based neural networks (CNN), multilayer perceptron (MLP), and transformer models. The model-centric methods focus more on the computational complexity of DNNs via efficient architectural approximations, compression, and efficient initialization for better generality. These methods focus on reducing the parameter size of DNNs while maintaining good performance. Specifically, architectural approximation focuses on adopting a simplified combination of operators to reduce the calculation costs in the training. It looks forward to exploring expressive alternatives of basic modules for general accelerations. Compression concerns efficiency in low-precision computing and sparse training, which are also required to be fully supported on the hardware implementation. Model initialization pays attention to searching for the better initial state with higher stability and generality, which could efficiently speed up the convergence and prevent the training process from crashing at the early stage. In summary, model-centric methods provide a promising approach to reducing the computational complexity of deep models for efficient training, which offers strong practicality and can be easily implemented in any deep learning framework.

    
    
    \item \textbf{Optimization-centric efficient training.} To improve optimization efficiency, we summarize three main factors, namely learning rate, batch size, and optimization objectives. The proper selection of learning rate and decay policy at different stages is a crucial issue in deep network training. However, it is challenging to find a universal method that works across different models and optimizers. Therefore, learning rate-centric methods aim to develop efficient and flexible strategies to train models efficiently and steadily. The second factor, batch size, also plays a critical role in optimization. With the parallel computing capability of GPU devices, increasing the number of samples in a single minibatch can improve training efficiency, especially when computing resources are sufficient. Hence, batch-size-centric methods usually focus on adopting large minibatch training to boost optimization speed. From the optimization perspective, we always strive to implement an objective with high stability, which is the primary concern of objective-centric methods. These methods focus on optimization objectives that provide generalizations that have robustness concerning data distribution and model architectures. In summary, optimization-centric methods study the efficient iterative calculations in the training process to provide the solid guarantees for the efficient training.

    \item \textbf{Budgeted efficient training.} Budgeted training is an approach that takes into account the available resources during practical training. It is mainly concerned with training efficiency in source-constrained scenarios, where computational resources such as training wall-clock time or calculation flops are limited. The primary objective of budgeted training is to ensure efficient and stable training while maximizing the potential of the model within the given constraints. This approach can lead to significant gains in the early stages of training. By adopting budgeted training, researchers and practitioners can make the most of available resources and avoid wasting them on inefficient models or training procedures. This approach can also facilitate the development of models that are more practical and suitable for real-world applications, where resources are often limited.

    \item \textbf{System-centric efficient training.} 
    System-centric methods focus on the practical implementation under the hardware support, which could turn algorithm design into real executable projects. Training large-scale models usually employs multi-node multi-device environments to implement parallel computing. It is primarily concerned with designing the underlying logic to address bottlenecks in communication across devices and coordinate the entire training process efficiently. Several open-source frameworks have been developed to significantly accelerate training deep networks. To efficiently utilize distributed training, the training process is distributed into smaller computational tasks that are executed on different nodes or devices in parallel. These nodes communicate with each other to exchange gradient updates and synchronize the overall training process. This distributed system enables training of large datasets and complex models that cannot be executed on a single machine. Several open-source distributed training frameworks have been developed, such as TensorFlow, PyTorch, and Horovod. These frameworks have enabled efficient distributed training on multi-node multi-device clusters and significantly reduced training time for large-scale deep learning models.    
\end{itemize}

In summary, our survey reviews the general training accelerations for efficient training. In the sections ``data-centric", ``model-centric", ``optimization-centric", and ``budget training", we mainly focus on the comprehensive studies from the perspectives of algorithm design and methodology, while in the section ``system-centric", we focus on the practical implementation from the perspectives of paradigm innovation and hardware support.
The main contributions of this survey are stated as follows: 
\begin{itemize}
    \item We review the general acceleration techniques for training large-scale models from the ``Data", ``Model", ``Optimization", ``Budgeted training" and ``System" perspectives to summarize their technical routes and implementation of each component, which contributes to providing solid guidelines for efficient training with task-free and model-free.
    \item We compare the strengths and weaknesses of each component in training acceleration and demonstrate their insights and interactions, which could inspire us to rethink the design of the high-efficiency paradigm for training large-scale deep learning models.
    \item We provide a comprehensive analysis of each technical route and its major challenges in the practical scenarios, which could provide guidance on their promising developments in the future.
\end{itemize}


\medskip 

The main structure of this survey is organized as follows. In section~\ref{sec:preliminary}, we introduce some preliminaries including the basic modules in different backbones and pretraining of the large-scale deep learning models, datasets, and the detailed notations adopted in this survey. From section~\ref{sec:data} to section~\ref{sec:budgeted_training}, we introduce the details of characteristics and properties from the perspective of ``Data-centric", ``Model-centric", ``Optimization-centric", ``Budgeted training", and  ``System-centric" based on the iteration formulation~(\ref{eq:gradient-based descent}), which contain their different technical routes for training acceleration. We also analyze and assess the strengths and drawbacks of each implementation. This novel taxonomy could provide a clear and comprehensive guideline on the existing approaches for efficient training. In section~\ref{sec:conclusion}, we discuss and summarize the techniques in this survey and suggest some promising research directions in the future.

\section{Preliminary}\label{sec:preliminary}

\begin{figure}[t]
\centering
\begin{minipage}{0.32\textwidth}
\centering
\includegraphics[width=0.99\textwidth]{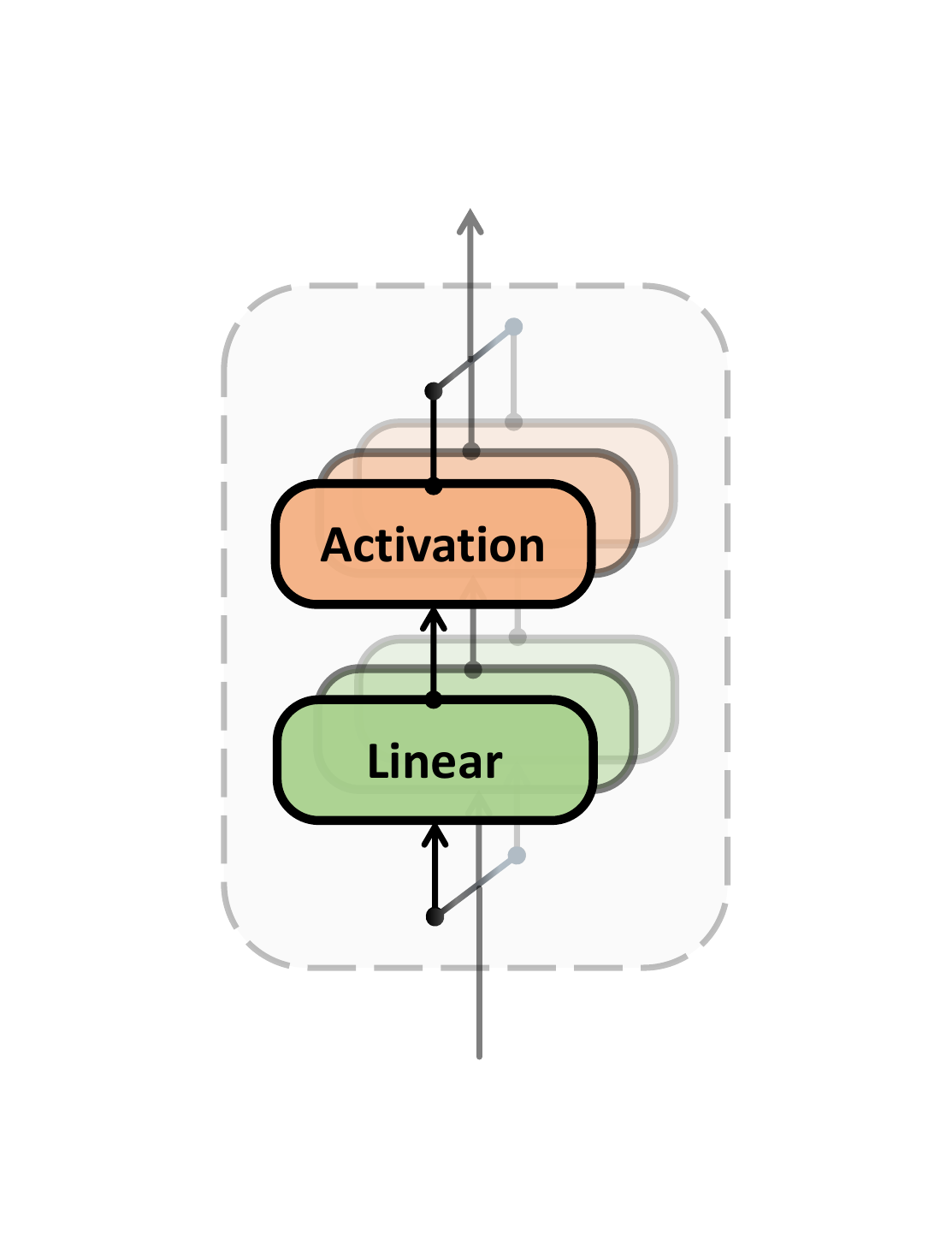}
\vspace{-0.4cm}
\caption{Basis of MLP.}\label{fig:mlp_module}
\end{minipage}
\begin{minipage}{0.32\textwidth}
\centering
\includegraphics[width=0.99\textwidth]{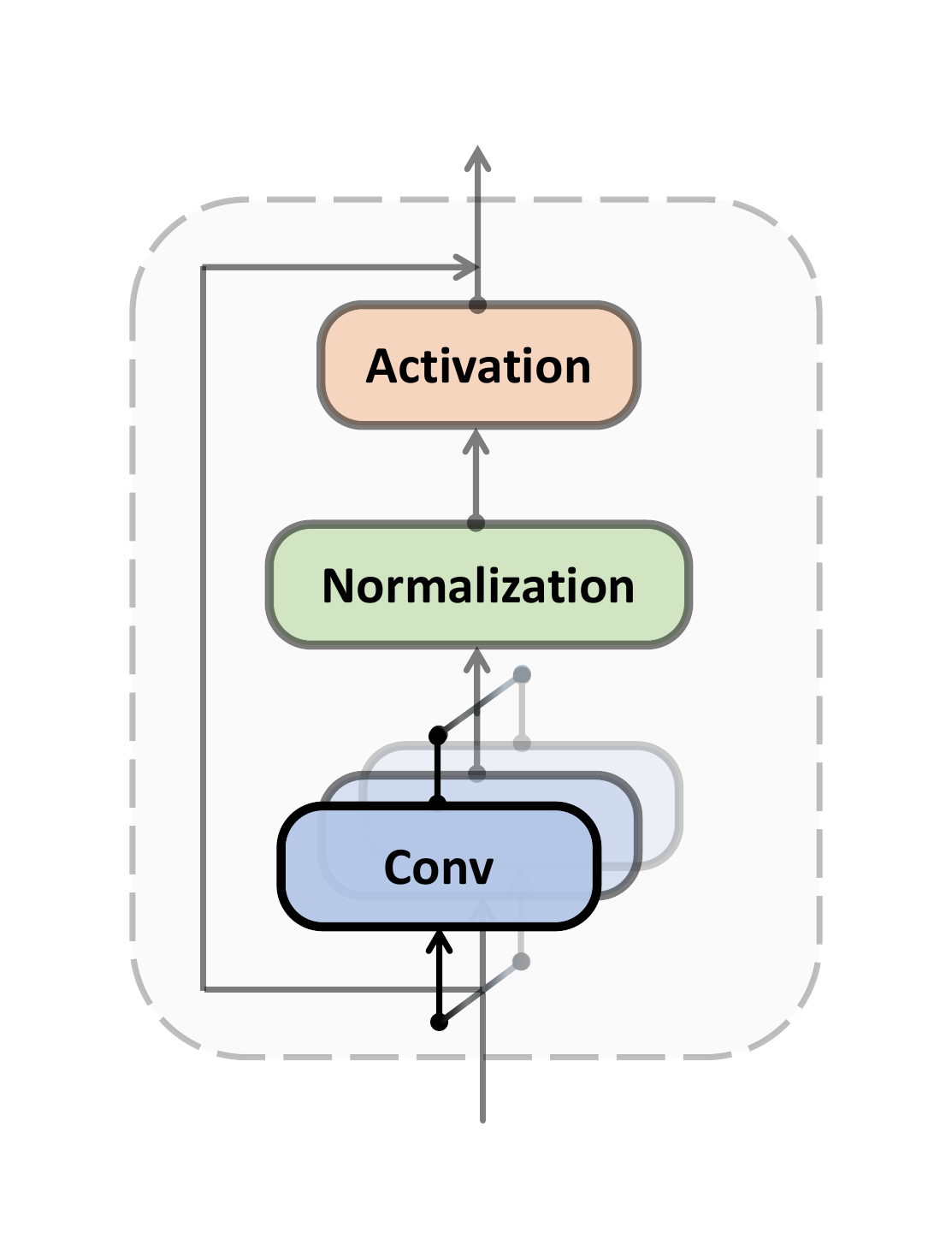}
\vspace{-0.4cm}
\caption{Basis of CNN.}\label{fig:cnn_module}
\end{minipage}
\begin{minipage}{0.32\textwidth}
\centering
\includegraphics[width=0.99\textwidth]{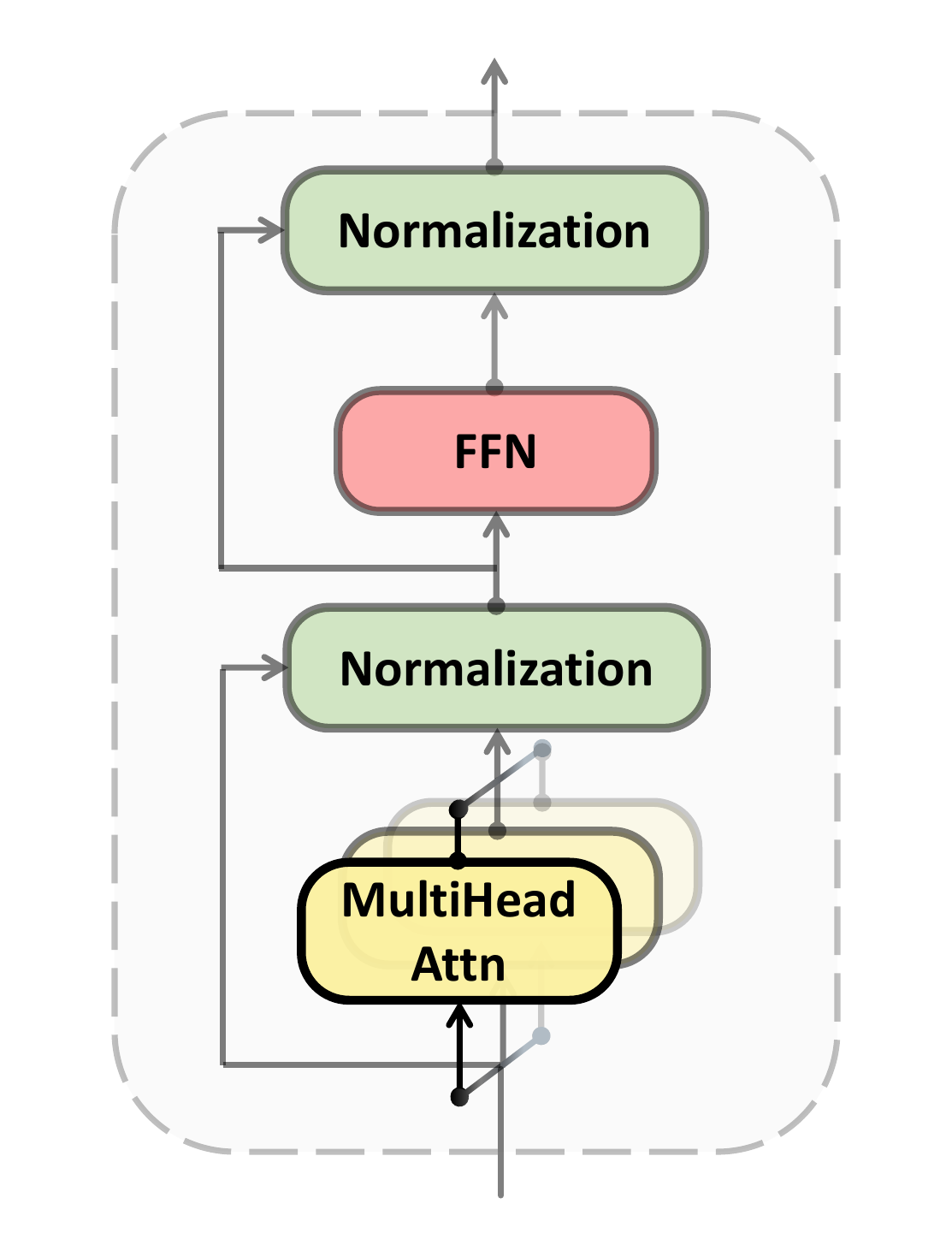}
\vspace{-0.4cm}
\caption{Basis of Transformer.}\label{fig:transformer_module}
\end{minipage}
\end{figure}

In this section, we first introduce several basic modules in large-scale deep learning models. Then we introduce some commonly used datasets and describe several notations in this survey.

\subsection{Basic Backbones}
In this survey, we focus on the efficient training of general deep learning backbones. Most of them are stacked by a sophisticated combination of the basic classical modules, mainly including MLP~(linear module), CNN~(convolution module), and Transformer~(multi-head attention module). In this part, we only refer to their general concepts. In section~\ref{sec:model}, we provide the specific formulation to illustrate how they work in the deep models and how they can be improved and optimized for efficient training.

\paragraph{MLP.} Multi-layer perceptron (MLP) is a widely used and classical model that has played a major role in the evolution of deep neural networks. MLP is composed of multiple linear layers, also known as the full-connection layer, with element-wise activation functions. During the training process, MLP generates numerous hidden neurons in addition to the input and output layers. Its parameters include a projection matrix and a bias vector. Figure~\ref{fig:mlp_module} shows a simplistic view of MLP, which demonstrates the combination of linear calculations with non-linear activations. Despite its simple structure, the performance of MLP can be enhanced by integrating modern network structures and training techniques. For instance, \textit{MLP-Mixer}, a fully MLP-based architecture proposed by \citet{tolstikhin2021mlp}, contains two types of MLP layers applied directly to or across patches, yielding state-of-the-art performance on image classification benchmarks.

\paragraph{CNN.} Convolutional Neural Network (CNN) is a powerful deep learning model that is extensively used for visual image analysis. CNN can be viewed as a multi-layer perceptron that is specifically designed to work with convolution. Its parameters include a kernel matrix, also known as the filter, and a bias vector. For every small area of an image, CNN operates with a similar set of kernels to extract high-level features, which can be considered as multiplying overlapping reconstructed images. Moreover, CNN ensures shift-invariance during the training process which adequately simulates the human visual system. Figure \ref{fig:cnn_module} depicts the fundamental building block of a CNN, which comprises a convolution layer, a normalization layer, and an element-wise activation. Some typical CNN structures include AlexNet~\citep{krizhevsky2017imagenet}, VGG~\citep{simonyan2014very}, Inception~\citep{szegedy2015going,ioffe2015batch,szegedy2016rethinking,szegedy2017inception}, and ResNet~\citep{he2016deep}.  
Moreover, CNNs have been optimized by adapting their architecture for more specific tasks. For example, Fully Convolutional Networks~(FCN)~\citep{long2015fully} is a variant of CNN specifically designed for semantic segmentation tasks. YOLO~\citep{redmon2016you} is an object detection network that achieves real-time performance by predicting bounding boxes and class probabilities directly from the input image. In summary, CNNs are an essential component of deep learning and have been evolving continuously to achieve state-of-the-art performance in various computer vision applications.

 
\paragraph{Transformer.} 
The vanilla Transformer~\citep{vaswani2017attention} is firstly proposed with an encoder-decoder architecture, designed for extracting information from natural language. The basic building block is called a cell, which is composed of two modules, Multi-head Attention (MHA), and a feed-forward network (FFN). MHA is a module that runs multiple independent self-attention layers in parallel to capture advanced semantics of inputs across various feature levels. This enables jointly attending to information from different representational subspaces and across different parts of the sequence. The FFN is a feature extractor that projects the advanced semantics from different MHA modules to the same feature space.
The Transformer architecture has also made significant contributions to the computer vision community. For instance, it has inspired the development of models like Vision Transformer~(ViT)~\citep{dosovitskiy2020image}, DeiT~\citep{touvron2021training}, and Swin Transformer~\citep{liu2021swin}. These models have achieved state-of-the-art performance in computer vision tasks, including object detection, image classification, and image captioning. Based on the Transformer architecture, there are various modified pre-training foundation models such as BERT~\citep{devlin2018bert} and GPT~\citep{radford2018improving}. The encoder-only BERT and decoder-only GPT have achieved remarkable results in NLP tasks such as question-answering, language inference, and text classification.
In summary, the Transformer architecture has contributed significantly to both the NLP and CV communities through its ability to capture and extract advanced semantics from inputs across various feature levels. Its development has facilitated the creation of novel models with state-of-the-art performance in a wide range of tasks.

\subsection{Large Scale Models and Dataset}

Both the increasing amount of parameters and growing dataset size make the training process of large-scale deep learning models even more time-consuming, which also requires that more advanced frameworks and systems should be developed to further accelerate the training process and reduce costs.  Below, we first introduce several commonly used large-scale deep learning models and the corresponding training datasets.

\paragraph{Large-Scale Models.}

Since \citet{vaswani2017attention} proposed Transformer in 2017, it has become the dominated model in NLP, leading to a boom of large language models (LLMs) and Vision Transformers (ViTs). 
Several (pre-trained) Transformer models demonstrate the development of efficient training, both in model structure improvements and in the application of efficient training techniques.
For NLP, there is a trend of rapid growth in the number of model parameters since GPT~(117M parameters)~\citep{radford2018improving}, the first pre-trained Transformer which is built of Transformer decoders and pre-trained unsupervised on BookCorpus~\citep{Zhu_2015_ICCV}.
Unlike the structure of GPT, BERT~(340M)~\citep{devlin2018bert} is composed of multiple layers of bidirectional Transformer encoders that can capture the relationship between positions. 
GPT-2~(1.5B)~\citep{radford2019language} has much more parameters than GPT-1 and is trained on larger datasets.
Using large-scale networks to train language models has proven to be a very effective strategy, and the growth of datasets, efficient model structures, and breakthroughs in hardware and software have made it possible, motivating models such as Turing-NLG~(17.2B)~\citep{microsoft2020turing}, GPT-3~(175B)~\citep{brown2020language}, Megatron-Turing NLG~(530B)~\citep{smith2022using} and Switch Transformer~(1.6T)~\citep{fedus2021switch} to further increase the number of parameters.
As for CV, ViT~(86M for ViT-Base)~\citep{dosovitskiy2020image} is a milestone in the application of Transformers in this field. Recently, inspired by the breakthroughs of large language models driven by scaling up Transformers, \citet{dehghani2023scaling} propose an efficient and stable training recipe that scales up ViT to 22B parameters (ViT-22B).
In the cross-modal or multi-modal area, CLIP~\citep{radford2021learning} turns the classification task into an image-text matching task by jointly training an image encoder and a text encoder, with ResNet-50~\citep{he2016deep} or ViT as the visual backbone and Transformer as the linguistic backbone. It is also released with DALL-E~\citep{ramesh2021zero}, a 12B parameter version of GPT-3 which is able to generate images from input text based on the pretraining of text-image pairs. 
Most of the above models are among or based on the \textit{Foundation Models} defined by \citet{bommasani2021opportunities}, i.e., models trained on a wide range of data generally using scaled self-supervision and can be adapted (e.g. finetuned) to a wide range of downstream tasks. 
With the proposed general-purpose large-scale model BEiT-3 by \citet{wang2022image}, they point out that the research on large-scale models in different domains is gradually approaching ``a big convergence": firstly, Transformers have become a common backbone architecture, secondly, generative pre-training has become the most important self-supervised learning method, and thirdly, scaling up data and model size further improves the performance of large-scale models. 
Although models are diverse, even the most gigantic models are composed of the basic modules of linear, convolution, and attention layers. For this reason, we summarize general accelerated methods and techniques for training large-scale deep learning models, learned from the mainstream pre-trained foundation models and some other work that is not limited to self-supervised training.

\paragraph{Datasets.}

Large language models require pre-training on vast datasets to learn common language representations for downstream tasks. 
For example, the BERT~\citep{devlin2018bert} model is trained on two large datasets, the BooksCorpus~(800M words)~\citep{Zhu_2015_ICCV} and English Wikipedia~(2,500M words).  
More recent and larger databases such as C4~(Colossal, Cleaned Crawl Corpus)~\citep{raffel2020exploring},  Pile~\citep{gao2020pile}, OSCAR~(Open Super-large Crawled Aggregated coRpus)~\citep{suarez2020monolingual}, mC4~(extended multilingual version of C4)~\citep{xue2020mt5}, BigScience ROOTS Corpus~(350B tokens)~\citep{laurenccon2022bigscience} are used by current models, such as LLaMA~\citet{touvron2023llama} that mixes resources from C4, GitHub, Wikipedia, Gutenberg~(a project containing public books) and Books3~(a section of the Pile), ArXiv, and Stack Exchange. 
Two of the largest multilingual datasets are OSCAR, which includes 152 languages and is 9.4TB in size as of January 2023, and mC4 which includes 101 languages and is 27T in size.
Datasets for selected language models from GPT-1~\citep{radford2018improving} to Gopher~\citep{rae2021scaling} are summarized by \citet{whatsinmyai}.

A large dataset may contain filtered web pages, books, articles, wikis, news, source code, and social media conversations. The model can learn based on a quantity- and quality-weighted sampling~\citep{brown2020language,touvron2023llama} of different parts of the dataset. 
In computer vision, ImageNet~\citep{deng2009imagenet} is a well-known large dataset of annotated photographs. 
Its most commonly used subset, ILSVRC~ (ImageNet Large Scale Visual Recognition Challenge)~\citep{russakovsky2015imagenet}, contains over 1 million images spanning 1,000 object classes.
As an example of the dataset for pre-trained models, the recent ViT-22B~\citep{dehghani2023scaling} is trained on a version of JFT~\citep{sun2017revisiting} that extends to approximately 4B images~\citep{zhai2022scaling}.



\paragraph{Notations.}

\begin{table}[t]
\centering
\caption{Notations used in this survey.}
\label{tab:notations}
\vskip 0.075cm
\begin{tabular}{cc}
\toprule
Components & Definition \\
\midrule
$A$ & matrix \\
$d$ & dimension of parameters where $w\in\mathbb{R}^d$ \\
$m$ & size of the dataset $(X,Y)$ \\
$H$ & input/output of the middle layer \\
$\phi$ & activation \\
$\Phi$ & normalization \\
$n$ & number of tokens \\
$h,b$ & height and width of a feature map \\
\midrule
$L$ & cross-entropy loss\\
$N$ & number of non-embedding model parameters \\
$D$ & dataset size in tokens \\ 
$C_{\min}$ & minimum computation to reach a given loss \\
\bottomrule
\end{tabular}
\end{table}

Table~\ref{tab:notations} shows some of the notation specifications used in this review. We use the capital letter to represent the matrix. $d$ is the size of the model. $m$ is the size of the dataset. $H$ is the feature map~(intermediate variable) in the training process. $\phi(\cdot)$ is the activation function and $\Phi(\cdot)$ is the normalization function. $n$ is the number of tokens in transformer models. $h,b$ are the height and weight of the input sample~(image). $L$ is the cross-entropy loss. $N$ is the number of non-embedding model parameters. $D$ is the dataset size in tokens as $D=nm$. $C_{\min}$ is the minimum computation to approach the target loss or target accuracy.
\section{Data-Centric Efficient Training}\label{sec:data}

Recent advances in large-scale models shine significantly while their requirements on the dataset increase dramatically. Enormous data samples are employed to drive the training process and achieve outstanding performance. Therefore, studies in the Data-centric are crucial for practical acceleration. The essential role of data processing is to efficiently increase the diversity of data samples without additional labeling. Data annotation is often too expensive to afford, which underscores the importance of research in the data-centric field. Meanwhile, it also focuses on improving the parallel loading efficiency of data samples.\\\\
In this section, we refer to all these efficient processing of data as ``data-centric" approaches that will significantly improve the performance of training large-scale models. We review and study techniques from the following perspectives:

\begin{itemize}
    \item \textbf{Data Regularization.} Data regularization is the pre-processing technique to enhance the diversity of raw data samples via a series of transformations. It improves the representation of training samples in the feature space equivalently without the requirements of additional labeling information. Efficient data regularization has been widely used in the training process and significantly improves the generalization performance of large-scale models.

    \item \textbf{Data Sampling.} Data sampling is an effective method to select a subset of samples from a large batch to perform the updates. It benefits from ignoring those unimportant or bad samples in the current batch to adopt the small batchsize training. Usually sampled data is more important, yielding a comparable performance as training with the full batch. The probability of each iteration is adjusted gradually with the training process to ensure unbiased sampling.

    \item \textbf{Data-centric Curriculum Learning.} Curriculum learning studies the progressive training setups at the different stages in the training process to reduce the total computation costs. In the beginning, training with the low-quality dataset is sufficient to learn the low-level features. Then adopting the high-quality dataset~(more augmentations and complicated pre-processing methods) gradually helps to learn the sophisticated features and achieve the same precision as the entire training.
\end{itemize}

\subsection{Data Regularization}

Training on a large, high-quality dataset will result in better performance than on a small, poorly organized dataset. For individual researchers, it is often impractical to collect and process large data manually, and they need to choose some open source datasets and consider combining multi-domain, multi-lingual datasets to improve data diversity and thus improve the generalization ability of the model. Therefore, corresponding pre-processing and cleaning methods such as de-duplication~\citep{lee2021deduplicating} (commonly with locality-sensitive hashing, LSH~\citep{shoeybi2019megatron}), removal of documents with few tokens for denoising, etc. may need to be applied. In addition, studying efficient dataset regularization helps to improve the training performance and generality of large-scale models. The purpose of data regularization is to reduce overfitting, make the model easier to adapt, and improve model generalization without affecting the training process too much. However, it usually introduces additional computational costs and sometimes becomes a bottleneck in the training process. For efficient training, it is crucial to select and test from a range of techniques and decide which one to use.\\\\
\citet{szegedy2016rethinking} propose label smoothing regularization which replaces labels by interpolating between the label distribution and another fixed distribution(e.g., the uniform distribution). This approach alleviates the problem of overfitting and makes the model more adaptive. \citet{he2019bag} show that label smoothing makes the distribution centered on theoretical values and with fewer extreme values. In addition, \citet{muller2019does} propose that label smoothing also improves the calibration of the model. They also observe that knowledge distillation into student networks is much less effective if a teacher network is trained using label smoothing. To explain these observations, they show by visualization that label smoothing encourages the representation of training instances to form tight clusters in the correct category but also leads to a loss of similarity information between samples from different categories, which negatively affects the effect of model distillation.

Data augmentation (DA) is another important regularization strategy for efficient training. By making better use of the original dataset, such as transformed images, data augmentation artificially and effectively increases the training data to improve the generalization ability of the model. Even some counter-intuitive methods of data augmentation are instead effective in improving the performance of the model. 
For computer vision, previous state-of-the-art models such as AlexNet~\citep{krizhevsky2017imagenet}, ResNet~\citep{he2016deep}, and EfficientNetV2~\citep{tan2021efficientnetv2} all use image augmentation techniques in training. 
There are many basic data augmentation methods, most of which are relatively simple, such as rotation, translation, shearing, flipping, cropping and resizing, adjusting contrast and brightness, adding noise, Gaussian blur, color jittering, 
format conversion (e.g. from RGB color model to HSV representation, which stands for Hue, Saturation, and Value), 
filtering, etc. In recent studies, basic erasing methods are proven to be effective, such as CutOut~\citep{devries2017improved} which randomly cuts a square out of the input during training, and Random Erasing~\citep{zhong2020random} which also erases a rectangle region of the image and fills it with random values or a mean value. 
Many data augmentation techniques are based on the idea of Mixup proposed by \citet{zhang2017mixup}, which mixes up the pairs of features and their corresponding labels. 
There is a class of Mixup extensions, for example, \citet{yun2019cutmix} propose CutMix which cuts out a small patch of image and replaces it with another one. 
More recent studies include SaliencyMix~\citep{uddin2020saliencymix}, 
Manifold Mixup~\citep{verma2019manifold}, 
StyleMix~\citep{hong2021stylemix}, 
TokenMix~\citep{liu2022tokenmix}, 
Co-mixup~\citep{kim2021co}, 
Supermix~\citep{dabouei2021supermix}, 
and TransMix~\citep{chen2022transmix}.

More advanced work investigates the combined use of various data augmentation algorithms and enters the field of automatic data augmentation, of which AutoAugment~\citep{cubuk2018autoaugment} is an important representative, which uses a search algorithm implemented as a controller RNN in reinforcement learning to find data-driven augmentation strategies. However, it brings a heavy computational cost, so it can only be applied to small data sets. 
Other advanced algorithms such as Population Based Augmentation (PBA)~\citep{ho2019population}, Fast AutoAugment~\citep{lim2019fast}, and Adversarial AutoAugment~\citep{zhang2019adversarial} further reduce the costs.
Random combinations can also be efficient, for example, \citet{hendrycks2019augmix} propose a data augmentation technique called AugMix, which randomly selects different data augmentation methods and then mixes the augmented images. 
\citet{cubuk2020randaugment} propose RandAugment which randomly selects a certain number of methods from several image augmentation methods, a general magnitude for transformations, and applies them sequentially for each sample. They investigate the influence of different data augmentation methods and also show that their method can scale the learned data augmentation strategies to larger datasets and models with a small cost and get better performance.


In the field of NLP, data augmentation is usually done before the training process in a relatively different way than for images. According to \citet{feng2021survey}, some of the data augmentation techniques are:
Rule-based techniques such as \textit{Easy Data Augmentation} which is summarized by \citet{wei2019eda}, include randomly inserting, swapping, deleting, and synonym replacement. Among them, the synonym replacement methods are such as \citet{zhang2015character} propose to randomly replace a word or a phrase with its synonyms ranked by semantic closeness, \citet{wang2015s} propose to create new instances by replacing a word with the results from searching for the k-nearest-neighbor (KNN) of the word in the embedding vocabulary. 
Example interpolation techniques that are pioneered by Mixup, for example, SeqMix~\citep{guo2020sequence}.
Model-based techniques such as Back translation~\citep{sennrich2015improving,edunov2018understanding}, which translate the text to another language and then translate it back.
As LLMs show great abilities in language understanding, some recent studies are using state-of-the-art models for data augmentation, for example, ChatAug~\citep{dai2023chataug} use ChatGPT as the data augmentation tool to rephrase input sentence into more additional sentences.

We note that advanced data augmentation methods improve the robustness but also bring an additional load, and they are usually better suited for models with more parameters, larger sizes, and longer training time than for smaller models. 
In particular, when bottlenecks exist in other parts of the training process and performing data augmentation computations does not affect the final time, data augmentation methods can be added to enhance the effect and improve the throughput rate.
\citet{steiner2021train} conduct an empirical study which shows that data augmentation (they used Mixup
and RandAugment
) and model regularization, together denoted as \textit{AugReg}, is effective for relatively small amounts of data.

\subsection{Data Sampling} 

During training, updates are performed on batches of samples, and all samples within a batch are treated equally. However, this can result in the training process spending more time on normal or less informative examples. To speed up training and avoid this issue, data sampling can be adopted, including importance sampling, designing novel objectives, etc., to achieve data efficiency.\\\\
Importance sampling is a Monte Carlo method that involves sampling from a different distribution to help evaluate the properties of a certain distribution and reduce variance. In stochastic gradient descent, importance sampling focuses on the samples that have the most significant impact on the model parameters, which reduces the variance of the gradient estimates. This approach is useful to reduce computational costs when it is fixed or when the batch size is increased. It is important to choose an appropriate importance sampling distribution that accurately represents the original data distribution to avoid introducing bias.
\citet{katharopoulos2018not} derive an upper bound for the gradient norm and propose an estimator of the variance reduction using importance sampling. 
\citet{katharopoulos2017biased} propose that the loss value can be used as a metric for importance sampling instead of gradient norm.
\citet{jiang2019accelerating} propose Selective Backpropagation which speeds up training by skipping the backward pass for training examples with low loss,  thus using fewer samples and reducing backpropagation computations. They show that this method converges faster than vanilla standard SGD. Sampled batches support greater data throughput to accelerate the training process.

\citet{chen2022mask} propose to generate a mask on the parameters to freeze the backpropagation of the inputs. It retains the forward graph, and the statistic characteristics of all batches for the layers such as normalization maintain the mapping of the total dataset. 
\citet{zhang2019autoassist} focus on the worse samples during the training. They utilize a selector to highlight the training samples in a large batch those have larger gradients as a valid approximation. Samples with small gradients are considered "good points" in this epoch and the network needs to pay extra attention to the "bad points".
\citet{mindermann2022prioritized} establishes a novel \textit{RHO} loss function to select a subset from a large batch and update the loss from the estimation of Bernoulli sampling. \textit{RHO} measures the importance of inputs with different loss values under its own evaluation yardstick, and then selects those important samples to update the models.
Similar ideas are implemented by \citet{kawaguchi2020ordered,ni2022k}. 
In \citep{ni2022k}, they propose the \textit{K-SAM} to select two subsets of samples on the gradient calculations which could efficiently reduce the backpropagation consumptions. 
\citet{kawaguchi2020ordered} study the \textit{Ordered-SGD} which directly selects the top-$q$ samples in a minibatch to maximize the loss function to perform the update in each iteration.
In each minibatch, $k$ samples are selected according to the loss value for training, so that the small minibatch can achieve the competitive training effect of the large minibatch. \citet{liu2022rsc} adopt the low rank and sparse inputs to accelerate the aggregation in the graph neural networks. They partition the input graph with several sparse blocks to reduce the computation of dense matrices with $\mathcal{O}(N^2)$ complexity. 
\citet{he2022accelerating} simulate the decentralized framework with edge devices of different computational abilities. They remove the slowest device in the training alternately to directly reduce the delay time costs, which can be thought of as a variant of the dynamic batchsize approach.
\citet{xie2023data} work in the reduced feature space to make importance weight estimation tractable over the space of text with the \textit{KL} reduction, which efficiently trains the language models with importance sampling.
\citet{sujit2022prioritizing} apply a similar idea to clip the input samples for reinforcement learning.
\citet{cai2022efficientvit} discard the dissimilar patches to achieve the same effect of reducing data in the version transformer models.

The acceleration methods based on the sampling focus on directly reducing the total amount of data involved in the calculation. These studies indirectly verify the data redundancy in the training. In other words, the ratio of the amount of data involved in training to the amount of data in the total dataset becomes one of the important indicators to measure the utilization efficiency of samples. How to design algorithms with high utilization and low computational costs will continue to be explored in the acceleration.

\subsection{Data-centric Curriculum Learning}

Curriculum learning~\citep{elman1993learning, bengio2009curriculum} is based on the intuition of starting with learning from simple tasks and gradually increasing the complexity of the tasks. This method has also been applied in various ways to pre-training and has shown good acceleration performance in a number of works. 
\citet{wu2021when} show that curriculum learning can improve performance when the training time budget is limited or when noisy data exists.  In applying curriculum learning, one of the more important challenges is finding a suitable criterion to judge the difficulty of the sample. \\\\
Progressive Resizing~\citep{karras2017progressive, howardl2018progressive} is a technique that trains small down-sampled to large full images with the training process to improve model performance and time to convergence. For example, a set of suggested parameters is that the initial image is scaled to half of the full image, with increments of 4 pixels at a time during half of the training process, and the last 20\% process is for fine-tuning.
\citet{touvron2019fixing} propose FixRes and show that training a network with a small resolution but inferring it with a higher resolution can result in a better generalization. Training with low-resolution images also helps to save training costs, when training with 224 and testing with 384 is better than training directly with 384. 
\citet{wang2022efficienttrain} introduce a clipping operation in the Fourier spectrum of the input image, which allows the model to learn from the low-frequency components first, as a form of curriculum learning, and therefore reduces training time.
\citet{koccyiugit2023accelerating} adopt the progressive resolution training~\citep{tan2021efficientnetv2}, 1-cyclic learning scheduling technique~\citep{smith2019super}, and hard augmentation selection technique for training self-supervised learning tasks and achieve training speedup. 
For NLP tasks, the short/long sequences correspond to the low/high-resolution images. \citet{press2021shortformer} show that training transformers with a short subsequence first and then switching to a longer subsequence can achieve speedup. \citet{ding2021progressive} find that training the machine translation model with a word-phrase-sentence data learning order could efficiently facilitate cross-lingual modeling, thus obtaining better translation. During the training process, we should keep the number of tokens and increase the batch size when low-resolution images are used.
\citet{li2021curriculum} propose the Sequence Length Warmup which linearly increases the sequence length during the early training process. In their experiments replicating the GPT-3 model (125M), they show that their method is able to train stably with a larger batch size and higher learning rate, and outperforms the original GPT-3 training recipe using fewer data and less time. 
\citet{li2022stability} propose a curriculum learning based adaptive length tuning technique to stabilize the training process of GPTs, which results in an efficient large-batch training schedule with wall-clock speedup. 
\citet{Nagatsuka2021PretrainingAB} propose to gradually increase the block size of input text for BERT training, showing faster convergence and higher training efficiency than a RoBERTa\citep{liu2019roberta}-baseline.  

\subsection{Discussion}

In this section, we review efficient training techniques from a data-centric perspective. We focus on the data efficiency of dataset regularization including the augmentation and pre-processing to enhance its diversity, sampling the valid subset from a large batch~(or entire dataset) to improve the training efficiency, and curriculum learning methods to alleviate the expensive consumption in the early stage of training.

Data regularization is an effective way to expand the diversity of data samples without additional labeling. Because of the complicated and random transformations, the dataset is equivalent to being expanded multiple times, which helps train the large-scale models with better generalization performance. However, overuse of data regularization introduces huge biases, resulting in poor quality. As more regularization methods are available, it is important to select effective combinations. In terms of data sampling, it can achieve the same or even better performance with small-batch training. Data sampling via well-designed algorithms can effectively increase the capacity of the training model. Usually, at the beginning of the training process, the model benefits more from the worse-performing samples. Ignoring good samples helps speed up the balance of learning the characteristics of different labeled samples. And, curriculum learning provides a progressive pipeline for processing the dataset. It allows feeding the low-resolution data with less augmentation to extract the coarse-grained features and gradually enhances the data quality to capture the fine-grained features. It achieves practical accelerations while maintaining high performance.

In the future, we think that promising studies from the data-centric perspective include:
\begin{itemize}
    \item Efficient regularization implementation. More task-specific approaches to data regularization are worth studying. Designing novel data regularization methods based on prior knowledge can further improve the generalized performance of large networks. Meanwhile, adopting the parallel processing of data regularization can also greatly accelerate the training process.
    \item Efficient data sampling. Novel sampling methods are expected which jointly consider the training acceleration and generalization guarantees~\citep{mindermann2022prioritized}.

\end{itemize}

\section{Model-Centric Efficient Training}\label{sec:model}

Designing efficient model architectures is always one of the most important research in the field of deep learning. An excellent model is an effective extractor and could be projected into high-level features that are easily separated. Different from other works with an extra focus on those efficient novel structures, our model-centric study pays more attention to the equality alternatives to the general modules, which achieve higher efficiency with comparable performance. Almost all the large models are made up of small modules or layers. Therefore, our review can provide good guidelines for efficiently training large-scale models.\\\\
In this section, we investigate those efficient ``model-centric" techniques from the perspectives as follows:

\begin{itemize}
    \item \textbf{Architecture Efficiency.} With the rapid increase of parameters in the deep models, it also introduces huge computational consumption. Therefore, implementing an efficient alternative to approximate the vanilla architecture becomes a trending topic. This substitution is not just an approximation of numerical calculations but also includes structural simplification and fusion in the deep models. In this part, we distinguish the existing acceleration techniques according to different architectures and demonstrate their insights.
    \item \textbf{Compression Training Efficiency.} Compression has always been one of the research directions of computational acceleration, which plays a key role in digital signal processing~(multimedia computing/image processing). Traditional compression consists of two main branches: quantization and sparsity. We will elaborate on their existing achievements and contributions to the deep training. 
    \item \textbf{Initialization Efficiency.} Initialization of model parameters is a very important factor in both existing theoretical analysis and practical scenarios. A bad initialization state can even cause the whole training to crash and stall at an early training stage, while a good initial state helps speed up the entire convergence within a smooth loss landscape. In this part, we mainly study the evaluation and algorithm design from the perspective of model initialization.
    \item \textbf{Model-centric Curriculum Learning.} From the model-centric perspective, curriculum learning usually begins to train a small model or partial parameters in the large-scale model and gradually recover it to the entire architecture. It shows the favorable capability in accelerating the training process without significant negative effects. In this part we review its implementation and efficiency in the training process.
\end{itemize}

\subsection{Architecture Efficiency}
The technique we review is to take efficient alternatives to accelerate calculations on the existing model modules. As the cornerstone of a large model, each basic module has different properties and should be studied individually. In this section, we mainly study the efficiency of the following modules, which are widely used in the construction of large models. Our review clearly points to the efficiency of general acceleration techniques from the model-centric perspective.

\subsubsection{Efficient Linear Layer} 

The linear layer is the basic module in the traditional models, which is also called the full-connection layer in the MLP. It could be considered as:
\begin{equation}
    H_{out}=HW+b,
\end{equation}
where $W$ is the weights and $b$ is the biases. It is the underlying matrix multiplication, so many fundamental optimizations on multiplications yield benefits to save resources and accelerate its calculation. Besides, in deep training, plenty of works study its approximation, which can reduce the parameters and calculation complexity in practice.
\citet{li2015butterfly} introduces a GPU-friendly butterfly factorization to split the multiplication as two recursive block diagonal matrices, which achieves comparable performance with only $1/r$ parameters~($r$ is the rank controller). Theoretically, it reduces a $N\times N$ matrix into the product of $\mathcal{O}(\log N)$ matrices with only $\mathcal{O}(N\log N)$ operation and memory required.
\citet{meng2022butterflyflow} propose the efficiently-computable and -invertible butterfly layers based on the butterfly factors, which contain a Jacobian determinant with less complexity. And, to expand this into the computing devices, they consolidate the architecture of the \textit{ButterflyFlow} as the independent module for training the deep models. 
\citet{song2022dnn} study a specific GPU-friendly sparsity on the MLP layer to accelerate the training. It adopts an approximate random dropout with a regular and online row-based or tile-based dropout. This sensitivity-aware dropout helps to reduce the required input feature maps in both forward and backward processes. Linear acceleration focuses on the low-rank and sparse matrix multiplications to maintain comparable performance. In the current large model, it is not dominant in the time consumption, and computation is relatively simple compared to other modules. Fundamental works could inspire us to design more efficient structures for accelerated training.

\subsubsection{Efficient Convolution}
Convolution is always an effective technology to extract high-level features for downstream tasks, especially in several computer vision tasks. It maintains the translational invariance and rotational invariance on the images which can efficiently capture detailed features. The general convolution calculations on the deep models can be written as:
\begin{equation}
    H_{out}(x,y) = \sum_{i=0}^{h-1}\sum_{j=0}^{b-1}H(x-i,y-j)W^c(i,j),
\end{equation}
where $(h,b)$ are the height and width of the input $H$~(we consider the single feature map), $W^c$ is the convolution kernel.\\

It can be seen that convolutional calculations are quite repetitive, where a large number of matrix multiplications make them time-consuming extremely. Therefore, acceleration on convolution is always an important research topic in deep networks.
\citet{tai2015convolutional} utilize the low-rank decomposition of tensors to evaluate the redundancy in the convolution filters and develop a new low-rank constraint on the parameters. It achieves a significant speed-up training and better performance than the vanilla convolution. Extensive empirical experiments verify its efficiency with comparable test accuracy.
\citet{zhao2018faster} propose a hybrid method that combines the Winograd minimal filtering and Strassen algorithm to dramatically reduce the computational complexity. It jointly reduces the required flops and the number of convolutions in the network simultaneously and achieves nearly no accuracy lost on the test of the visual geometry group networks~(a structure of pure convolution).
\citet{9027244} pay more attention to the efficient implementation of embedded CPU devices with limited memory and budget resources. It fills the gaps and transfers the neural networks in mobile devices to support local applications.
\citet{8966288} study a novel fast convolution algorithm to remove the redundant multiplication operations at the controlled units on both the 1D and 2D convolutions. It also processes the input features and generates the outputs via a flexible block size which is independent of the kernel size. And, the memory access efficiency is also largely improved by the alternate computing neurons.
\citet{wang2022one} present a new approach that study the 1D convolution in the deep networks, named \textit{ODLS}~(one-dimensional deep low-rank and sparse). It simplifies the architectures and makes the deep networks much easier to be trained and generalized in accelerated magnetic resonance imaging tasks.
\citet{ghimire2022survey} summarize the mainstream techniques of accelerating convolution computing and hardware-efficient implementation, including decomposition, compression, low-rank approximation, and sparsity. Though transformer models achieve the SOTA results and impact the use of convolution layers, some of the current efficient architectures consider combining the transformer and convolution modules to further extract the detailed features for the computer vision tasks~\citep{li2021localvit, xu2021vitae}. In the future, convolution will still play a huge role in architectural design. As an efficient feature extractor, accelerating its implementation is also an important topic to explore.





\subsubsection{Efficient Activation} 
The activation function is a handcraft function to element-wise scale the inputs to help the deep models learn the complex patterns and high-level features. Similar to a neuron-based mechanism in our brain, it ultimately determines what gets fired to the next neuron. It could be represented as:
\begin{equation}
    H_{out} = \phi(H),
\end{equation}
usually $\phi(\cdot)$ is a non-linear function.\\

Activation functions create nonlinear maps for deep networks which effectively improves the potentiality to fit the complicated objective functions in the scenarios. Since \textit{ReLU}~\citep{hahnloser2000digital,krizhevsky2017imagenet} is proposed to simulate the bionics, deep neural networks have made great strides in many fields. It keeps gradients from back-flowing from layer to layer without dissipating or exploding. A series of related optimization activation functions are also applied, such as \textit{LReLU}, \textit{ALReLU}, and \textit{LeLeLU}. As huge models are progressively designed to solve large tasks, more efficient activation functions are implemented.
\citet{clevert2015fast} propose the \textit{ELU} activation function which can be considered as a \textit{Tanh} with two-side saturation. It results in a negative output ruled by the exponent multiplication to solve the vanishing gradients and dead state problems to accelerate the training process.
\citet{hendrycks2016gaussian} study the \textit{GeLU} to weight inputs by their specific values instead of signs. It contributes to improving the flatness of various large models. Smoothness in the early stages of training effectively reduces the number of training epochs and thus speeds up.
To further promote the stability of the data flow in the network, combining techniques from data normalization, \citet{klambauer2017self} develop the \textit{SeLU} activation to be able to make the output into a normal distribution across all layers. The uniformity of the outputs helps to maintain valid and effective gradients in backpropagation and to learn faster.
As regards the benefit of the shifting means around zero value, \citet{carlile2017improving} design the \textit{ISRLU} to save the high computation costs of the exponent function on the left side of zero. The replacement of polynomial-like complexity is much lower than the exponent function and achieves efficient training in MLP networks.
\citet{raffel2020exploring} verify the efficiency of the \textit{SwiGLU} activation in the FFN modules, which is an important component of the transformer. The finetuning coefficient can improve the distribution similarity of outputs across different layers. It tends to require lower training epochs in pre-training tasks. 
The calculation of the activation function is usually element-wise and accelerated by the parallel computing units. In the practical forward and backpropagation, the time costs of these layers are generally narrow (lower than $10\%$). Appropriate activation functions can provide higher smoothness to the network on the different stages, enabling a faster transition to areas with smaller loss values during training.

\subsubsection{Efficient Attention}
Attention has successfully developed the transformer models and achieved SOTA results on several tasks. The current general attention module is the multi-head attention~(MHA), which employs many self-attention branches to focus on the different traits. We denote the $H$ as the inputs vector, and the whole inference is divided into two parts,
\begin{equation}
    Q_i, K_i, V_i = HW_i^Q, HW_i^K, HW_i^V,
\end{equation}
which are the projected queries~($Q$), keys~($K$), and values~($V$), respectively. 
Then the self-attention calculation performs as follows:
\begin{equation}
    H_{out} = softmax\left(\frac{Q_i K_i^\top}{\sqrt{d}}\right)\cdot V_i,
\end{equation}
where $d$ is the dimension. In each head, each token will calculate an inner product with the others to generate the relationship value, and then normalize to a ratio within the range of $\left[0,1\right]$ to equitably calculate the intensity of attention. Therefore, if the number of the tokens is $n$, it requires $\mathcal{O}(n^2)$ computing complexity, which also makes self-attention modules mostly time-consuming.\\

Before the introduction of the transformer architectures, the attention mechanism is implemented by complicated RNN-based encoder-decoder modules with the $\mathcal{O}(n)$ complexity and sequential operation. Transformer models~\citep{vaswani2017attention} study the alternative scaled-dot product modules to free the recurrence calculations and achieves a similar effect. Computational issues are less important on small tasks~\citep{geiping2022cramming}. When the dimensionality of the data increases significantly, the self-attention module will play a key role and account for the dominant costs of training the transformer models. Therefore, a series of works are carried out around how to efficiently calculate the attention module. 
\citet{kitaev2020reformer} adopt locality-sensitive hashing with the reversible residual layers instead of the vanilla dot-product attention, which reduces the complexity from $\mathcal{O}(n^2)$ to $\mathcal{O}(n\log n)$ in the \textit{Reformer} model.
\citet{choromanski2020rethinking} estimate the full-rank-attention under the provable \textit{Performer} by the positive orthogonal random features approach to maintain the linear space and time complexity without relying on any priors.
\citet{wang2020linformer} propose the \textit{Linformer} to approximate the self-attention by a low-rank matrix with $\mathcal{O}(n)$ complexity in both time and space.
\citet{xiong2021nystromformer} study the linearized self-attention approximation via Nystr$\rm\ddot{o}$m method in the \textit{Nystr$\ddot{o}$mformer}.
\citet{peng2021random} put forward a linear time and space attention \textit{RFA} based on the random feature methods to offer a straightforward solution through an optional gating.
\citet{pan2021ia} focus on the version tasks and propose the novel $\textit{IA-RED}^2$ framework to remove the uncorrelated tokens at different stages, resulting in the effective shrinkage on the inference calculations.
\citet{ma2021luna} utilizes an additional fixed sequence as inputs, which allows the $Luna$ framework to perform attention operations linearly with storing adequate contextual information.
\citet{wu2021fastformer} consider each token representation with global context instead of modeling the pair-wise interaction among the whole tokens and realize the efficient \textit{Fastformer} under the theoretical linear complexity. 
\citet{ren2021combiner} perform the full capability in each attention head while maintaining lower computation and memory complexity from the drop-in replacement of treating the self-attention module as a conditional expectation over each location embedding.
\citet{chen2021scatterbrain} incorporate the classical robust-PCA to the attention matrix multiplication to unify the sparse and low-rank calculations.
\citet{rabe2021self} provide a practical and stable implementation for accelerating the self-attention modules into $\mathcal{O}(\sqrt{n})$ memory costs.
\citet{liu2022ecoformer} customize to high-dimensional attention via kernelized hashing to align the vanilla query and key vectors into low-dimensional binary information in Hamming space by the \textit{EcoFormer} paradigm, which is learned from the ground-truth similarity relations.
\citet{li2022accelerating} formulate a threshold to select the most pertinent tokens trained by a soft differentiable regularizer integrated into the global objective function, which efficiently reduces the calculations and maintains the same test error.
\citet{pan2022fast} introduce the \textit{HiLo} attention, which distinguishes the image information with the frequencies, to separate the large token cluster into different small groups and perform the group-wised self-attention encoder.
\citet{zhang2022mixhead} re-combine the hybrid multi-head attention mechanism to integrate the shared information of each token across the different heads, which is similar to the group division methods on the head dimension.
\citet{hua2022transformer} develop the gated attention unit to allow the use of a weaker single-head attention module with quality loss and then provide a linear approximation to this novel variant.
\citet{qin2022devil} summarize the major bottlenecks in the previous linear-designed attention modules as unbounded gradients and attention dilution, which cause the performance gaps. To further address these issues, more attention should be paid to the scaling of attention matrices.
\citet{dao2022flashattention} consider the co-designed methods via the hardware and propose the \textit{FlashAttention} implementation to practically accelerate the self-attention calculations on the wall-clock time, which has been assembled in many ML frameworks, e.g. TensorFlow, and PyTorch.
A multi-scale visualization of attention in the transformer model gives a visualization of different heads and analyzes MHA that some can be pruned except for some specialized heads. 
In future research, how to implement the attention mechanism efficiently and equivalently is still the main direction of exploration, especially in the face of those tasks with huge models and massive amounts of data samples.
As a comprehensive review in the field of high-efficiency transformers, the work of \citet{tay2022efficient} presents the summary and categorization of linear attention methods.





\subsubsection{Efficient Normalization}
Normalization is an important technique to attenuate the covariance shifts in the models. When the initial state is bad, the beginning stage of the training will be affected significantly by the imbalanced features, which causes the instability of the gradients. Therefore, normalization is introduced to regularize the mean and variance of features. Both convolutional neural network~(CNN) and transformer are beneficial from the unified dynamics. We can note it as:
\begin{equation}
    \Phi(H) = \left(\frac{H-\mu_H}{\sigma_H}\right)\odot W + b,
\end{equation}
where the $W$ and $b$ are the weights and bias respectively, and $\mu_H$ and $\sigma_H$ is the statistical mean and variance of the inputs $H$. Their statistical calculations are determined by the normalization methods.\\\\
Traditional normalization employs the \textit{BN}~(batch normalization) to avoid the gradient diminishing in the backpropagation and it makes great progress in several CV tasks. It could balance the mean and variance of the samples in a batch and maintain stability in the training. Due to the batchsize becoming small in the NLP tasks, the \textit{LN}~(layer normalization) is adopted. In the current SOTA transformer models, it is still inherited this technique. Besides these, there are \textit{IN}~(instance normalization) and \textit{GN}~(group normalization). They differ from each other by which dimension to normalize the data samples.
\cite{hoffer2017train} propose the \textit{GhostNorm}~(ghost batch normalization), a batch normalization technique that divides the batch into smaller batches and normalizes independently them in parallel. This technique reduces the required pin memory in each communication and makes full use of the buffer cache.
\cite{dimitriou2022sequential} propose \textit{SeqNorm}~(sequential normalization) to normalize the inputs across two dimensions of the input feature maps. It tactfully combines the \textit{GhostNorm} and \textit{BatchNorm} and outperforms the others on the classification tasks.
\citet{zhang2019root} expand the layer normalization to the \textit{RMSNorm}~(root mean square layer normalization) which assumes that re-centering invariance in the LN is dispensable and regularizes the summed inputs via their root mean square. It is computationally simpler and more efficient than vanilla LN.
When it is employed in the distributed system, the mean and variance calculations are actually the reducing operation, which is restricted by the communication bottleneck between the device memory and register. To save the limited computing resources, \textit{FusedLN} is proposed to merge several reduce operations in one single function, which can fully improve the GPU utilization with less memory allocated.
\citet{singh2022feature} study a novel \textit{FWN}~(feature wise normalization) approach to achieve the fast implementation yielding optimal performance. It benefits from the collective reply to normalize the input features as an individual unit to select a better combination, which could be optimized.
\citet{xia2022cross} propose the \textit{CoN}~(collaborative normalization) to eliminate the domain discrepancy and accelerate the training process for the unsupervised domain adaptation tasks. It measures the difference in the features from different domains and learns the specific statistics to excavate similar knowledge and representation. It improves the generalization accuracy and is easy to be plugged into the deep models without extra consumption.

\subsection{Compression Training Efficiency}

Compression is also one of the most mainstream acceleration methods at present. From the time consumption of the whole training process, most of the time is spent on inference~(about $30\%$) and backpropagation~(about $60\%$). Improving computing efficiency can significantly accelerate the entire training process. One of the efficient ways is quantization. Reducing the number of bits occupied by the raw data saves the inference time on the computing devices. Another way is sparse training. A large number of studies have shown that large neural networks have redundancy in parameters and gradients in the training process, which means that the update frequency on several parameters can be appropriately reduced. In this part, we focus on those which will maintain the total number of parameters in the whole training process. 

\subsubsection{Quantization Efficiency}\label{sssec:quantization}

The basic and fundamental rationale for quantization to accelerate the training is that the deep networks would consume less memory and computations if the parameters and inputs are converted to low-precision storage. However, it is always a trade-off between quantization and accuracy. A series of studies have investigated the link between quantization and accuracy, and propose how to preserve training accuracy to the greatest extent in practice.

At present, the common algorithms use the standard FP32 floating point type for the inference and backpropagation of the parameters by default. This data type can meet the high precision requirements in the current deep training, especially on large dataset and huge models. Correspondingly, the half-precision data type FP16 maintains the 5 bits exponent and 10 bits fraction and becomes a substitute for the FP32 in current training. In some evaluations, it can even approach the results of the single-precision-based calculations in less than half the vanilla training time. However, limited precision often faces the gradient overflow, especially in the early stages of training, which causes the core dump with NaN. And, low-precision calculations have a greater impact on the later stages of the training process. To balance the efficiency and accuracy, \textit{AMP}~(auto mixed precision) is proposed to be widely used in deep training. It utilizes empiricism to use FP16 for some unimportant parameters and use FP32 for other parameters to achieve the best performance, which is currently one of the most efficient acceleration techniques.
Based on this, \citet{wang2019haq} propose the \textit{HAQ} framework to leverage reinforcement learning to automatically determine the quantization policy on the hardware accelerators. It is fully automated and selects the quantization policy for different modules in the networks. 
\citet{huang2022hardware} introduce the guidelines to select
the hardware-friendly design options for the chip architects in deep models, e.g. the properly chosen quantization approach and the optimal mapping strategy in the practical scenarios.
\citet{ryu2022bitblade} propose an efficient precision scalable accelerating structure with novel bit-wise summation and channel-wise aligning scheme. It significantly reduces the area overhead for the scaling and correctly fetches inputs and parameters from the on-chip SRAM buffers.
\citet{dettmers20218} try the block-wise quantization with the 8-bit optimizer which could achieve comparable performance. Block-wise quantization splits inputs into several blocks which are independently quantized in parallel, yielding faster optimization and higher precision quantization. This is a straightforward way to simplify the calculation. Realizing quantization methods, which can be effectively supported on the computation of hardware devices, is still one of the most important research directions in the future.

\subsubsection{Sparse Efficiency}\label{sssec:sparsity}

Sparse computing requires the support of the hardware, otherwise, the acceleration cannot be achieved in the practical scenario. 
\citet{brock2017freezeout} propose to freeze the early layers progressively and exclude them from the backpropagation during the training process. Through experiments on CIFAR, They empirically show that their algorithm \textit{FreezeOut} results in a 20\% speedup for DenseNet~(3\% loss in accuracy) or ResNet~(without loss of accuracy), and no improvement for VGG networks. A similar idea can be referred to \citet{raghu2017svcca}. \citet{xiao2019fast} propose a method that intelligently freezes layers by calculating normalized gradient differences to decide the layer freezing rate. This implementation shows high efficiency at the mid-stage of the training.
\citet{liu2022communication} propose a novel gradient compression method~\textit{JOINTSPAR} for the large batch optimization via the layerwise adaptive learning rates. It drops the coordinates of the vanilla gradients by minimizing the combinations of gradient norms and parameter selection from these thresholds.
\citet{ding2022re} propose an effective method for convolutional networks to achieve reparameterization on the optimizer. It combines the convolution parameters between different branches in the parallel residual structure to achieve equivalent training. The reparameterized structure can compress the wider network into a narrow one while retaining the characteristics of the vanilla network.
\citet{kong2022peeling} point out the redundancy of the million-scale training dataset and propose an end-to-end practical framework \textit{Tri-Level E-ViT}. It applies both the importance of the data samples and sparse connection on the patches in an image to accelerate the ViT training.
\citet{huo2018decoupled} decouple the backpropagation using the delayed gradients, and remove the backward locking via splitting the networks into multiple modules. This decoupled parallel mechanism save the waiting time on the chain of the gradient backpropagation.
\citet{goli2020resprop} observe that over $90\%$ of gradients can be reusable during the training process. Leveraging this phenomenon, they propose the \textit{ReSprop} method to specify the gradient vectors by enabling the reduction in the backpropagation computations. To further match the fine-grained calculation of GPU devices, \textit{ReSprop} introduces a generic sparse convolution accelerator.
\citet{lym2019prunetrain} propose the \textit{PruneTrain} to gradually reduce the training cost. It adopts a structured group-lasso regularization approach that constrains the parameters to be small valued. It introduces the reconfiguration technique to implement the local pruning and reduce the computation in both FLOPs and memory.
\citet{nguyen2022improving} pay attention to the redundancy of the multi-head attention. They propose a mixture of Gaussian keys (\textit{MGK}) to replace the vanilla multi-head architecture. 
\citet{dun2022resist} consider the blockwise training methods and propose the \textit{ResIST} to accelerate the ResNet. In each round, it trains a randomly generated sub-network and communicates a small portion of the parameters, and merges the results together at the end stage.
\citet{fan2022adaptable} use the sparse butterfly multiplication accelerator to save the computational complexity which is supported by the hardware implementation. This type of matrix can be easily adopted on both attention and FFN modules and could save plenty of training wall-clock time. Due to the characteristics of butterfly matrices, the performance drops extremely lightly.
\citet{han2022turbo} propose the \textit{Turbo} training for the transformer models on the action classification tasks and video-language representation learning. They allow to drop off the tokens in the training to specify the input vectors and enable to split the long-schedule video frames into fragmented groups.
\citet{li2022efficient} open the \textit{Magicube} library which serves for the sparse matrices calculation on the Tensor cores. Each value can be represented as a low-precision integer on the A100 devices. It supports SpMM~(sparse matrix multiplication) and SDDMM to accelerate the sparse kernels without accuracy dropping on the sparse transformer inference.
\citet{bolya2022token} are inspired by the idea of sparse inputs and introduce \textit{ToMe}~(token merging) technique to implement the sparsification of the inputs. Different from directly removing the tokens, \textit{ToMe} allows the selection of several tokens and then merges the others into one token as the input vector for the transformer models.

In summary, the main approaches to the sparse training for acceleration focus on the following aspects:

\noindent
\textbf{Sparse Gradients.} The gradients determine the offsets on the parameters in each iteration. For parameters with small gradient values in an update, their offsets are also small. Compared with other parameters, pausing their updates will not hurt the training performance. Therefore, reducing the calculation of backpropagation by updating the sparse gradients can effectively accelerate the training.

\noindent
\textbf{Sparse Parameters.} Parameter redundancy is one of the very common issues in deep models. Many studies point out that on the different data samples, selective updates are often more generalized. Therefore, pruning and reconfiguration -based techniques to freeze partial parameters in the training could greatly save the time and memory costs of both forward and backward processes.

\noindent
\textbf{Sparse Inputs.} There is a large redundancy in the input samples. Usually, in a mini-batch of data, different categories of samples are trained to different degrees. Dimensionality reduction in the batch dimension can be achieved through some unique sampling methods, e.g. importance sampling. This allows supporting the larger batchsize to accelerate the entire training process.

\noindent
\textbf{Sparse Features.} Features are the function mapping of the input samples in the middle layers and they occupy plenty of computing and memory consumption. There is also great redundancy in the features. Aggregating the features of the middle layer through aggregation or extra attention mechanisms can effectively accelerate the training on the deep networks.

\subsection{Initialization Efficiency}
Studies on model initialization indicate that arbitrary initialization strategies may lead to a slow convergence or even completely stall at the beginning of the training~\citep{mishkin2015all}, especially on large-scale models. Therefore, proper initialization could do more with less on efficient training. Several previous studies reveal the relationship between initialization and training from different perspectives and propose a series of efficient methods to better initialize the models. Recent advances in pretraining techniques also indicate that a better initial state could help to understand the general knowledge of task-free and improve the performance of the downstream tasks. In this part, we mainly focus on the efficient initialization methods and the pretraining techniques. 

\paragraph{Train-from-scratch Initialization.} Model initialization plays an important role in efficiently training a model from scratch.
\citet{sutskever2013importance} find that both the model initialization and the momentum update are important in the training. They indicate that if the model starts from a poorly initialized state, the training process may fail with momentum updates. 
Previous studies demonstrate that the failure caused by the random initialization may come from the shrinking variance of the features in the deep layers. To tackle this difficulty,
\citet{saxe2013exact} propose that the proper scaling for the initial condition of the parameters matrix $W$ between two linear layers corresponds to selecting the initial parameters from the zero-mean and $1/d$-variance Gaussian distribution under an orthonormal matrix, where $d$ is the dimension of the matrix $W$. This initialization could maintain the norm of the inputs even in the deep models, which efficiently avoids the negative impacts of unstable gradients in the training.
\citet{sussillo2014random} show that a proper scaling on the random matrix of each layer leads to an unexpected shift on the log of the norm of the parameters, whose variance grows with the depth of the model. They propose to increase the width of the model and adopt an additional correction scalar to adjust the model initialization.
\citet{kumar2017weight} study the empirical evidence on the impacts of activation layers. The functions, like hyperbolic tangent and sigmoid, which are differentiable at zero, will show more sensitive properties than those that are not differentiable at zero on the random initialization. 
\citet{zhang2019residual} explore the specific initialization for the residual networks with the skip-connections.
\citet{huang2020improving} expand the weight initialization to the Transformer architectures with the guarantees of theoretical corrections. It allows training the model without the warm-up and layer normalization.
Learnable initialization also shows great potential in model initialization.
\citet{dauphin2019metainit} propose the \textit{MetaInit} to search for the good state with an approximately linear landscape. It adopts several gradient descents to tune the norms of the parameters via minimizing the curvature information of second-order effects.
\citet{zhu2021gradinit} find a better initial state for minimizing the loss function via several classical data samples.
\citet{zhao2021zero} propose the \textit{Zero-Initialization} to initialize the model with only zeros or ones. It adopts the Hadamard transformation to remove the whole randomness in the initialization which could be expanded easily to the sparse training.
Efficient initialization shows strong performance and potential in large-scale model training. It also inspires us to think about finding a trained model to further fine-tune parameters on the complicated downstream tasks, which is the pretraining technique.

\paragraph{Pretraining-finetuning Initialization.} Pretraining-finetuning mode is a efficient paradigm to achieve the excellent performance on the downstream tasks. It comes from the transfer learning which allows training the large-scale model on an extensive dataset to learn a good feature extractor. Then, it freezes the backbones and finetunes partial parameters to achieve the good performance. The pretraining phase could be considered as searching for a better initialized state for the downstream tasks. Recent advances shine a dazzling light on pre-training on the pretraining. 
\citet{radford2018improving} propose a semi-supervised approach for the language tasks with adopting the initialization state trained via the unsupervised pretraining. It shows that this generative model performs better than the supervised training.
\citet{radford2019language} greatly increases the model parameters and data samples in the pretraining phase with joint multi-tasks. They indicate that under the sufficient data samples, the large-scale model has great potential for efficient training on the downstream tasks.
Furthermore, \citet{brown2020language} use a extremely large foundation model as the pretraining to test on the other tasks. It could achieve SOTA performance on some few-shot and even zero-shot scenarios. 
\citet{he2022masked} also use a similar training strategy to pretrain a CV model, which makes a great success on the downstream tasks in the CV community.
Pretraining provides an excellent initialization of the model for efficient training. Current trends prefer a combination of large-scale model and extremely extensive data samples. Therefore, how to implement the efficient training on the pretraining is a promising research direction in the future.

\subsection{Model-centric Curriculum Learning}
Model-centric curriculum learning mainly concerns on circumventing the training difficulties in the large-scale models. It explores the small capacity for the large-scale models at the early stage of training and suggests to reduce the number of parameters involved in the calculation accordingly. With the training, the capacity of the model increases gradually and all the parameters are recovered to complete the entire training process. It shows great potential in efficient training, especially on large-scale models.\\\\
\citet{bengio2006greedy} propose a greedy approach to achieve progressive layer-wise learning. This inspires the potential of progressive training on the model-level.
\citet{karras2017progressive} study the curriculum learning in the training GANs. It progressively increase the layers of generator and discriminator in synchrony. The new layers are added to the training smoothly to avoid the sudden shocks on the well-trained layers.
\citet{smith2016gradual} propose the \textit{DropIn} to train the deep models from a shallow sub-net and then gradually add the new layers to maintain its efficiency. They demonstrate that model-centric progressive training works as regularization term on the parameters.
\citet{wang2017deep} focus on the semi-supervised learning and design the deep growing learning. They formulate the training as a EM-like process and enable the large-scale models alternately iterates between automatically growing layers. They indicate that progressive training may help the large-scale models maintain higher stability and generality.
\citet{gong2019efficient} study the unsupervised pre-training techniques and propose the \textit{stacking} algorithm to transfer the fundamental knowledge from the well-trained shallow model to a enlarged model. In the transferring, a progressively increased stacking is adopted to accelerate the training of BERT.
Similar idea is proposed by \citet{li2020shallow} on the neural machine translation~(NMT) task.
\citet{zhang2020accelerating} learn a progressive schedule to add extra-stableness for the trained model with layer dropping. A global list of dropping rate across all the ST blocks are set and updated for different layers in the training.
\citet{gu2020transformer} observe that it is beneficial to adopt the joint growth on multiple dimensions of the models. They propose to explore the alternative growth in each dimension instead of only the layers along the depth.
\citet{li2022automated} relax the optimization problem of the progressive learning to a sub-network optimization process, which could be solved by a automated one-shot estimation. It also minimizes the searching overhead by recycling the parameters of the sub-nets.
Model-centric curriculum learning achieves great success in large-scale model training. In the future, designing the sophisticated progressive schedules to match the data samples could be a promising study.









\subsection{Discussion}

In this section, We focus on how to achieve efficient training acceleration in a specific large-scale model. We consider general approaches to accelerating from the perspective of the essential training process, including adopting alternative efficient architectures to reduce the computational complexity, compressing the tensors participating in the training procedures to consume less memory and computational costs, and efficient model initialization methods for the fast convergence.

In terms of the model architecture, our focus is to review the efficient alternatives of the fundamental modules instead of exploration on designing efficient models, including the basic components linear, convolution, attention, activation, and normalization. Simple structures usually tend to be more sensitive to parameters. The performance of fundamental modules like linear layers and convolution layers is easily affected when adopting their low-parametric alternatives. This limits the cap on the practical acceleration ratios. Utilizing efficient alternative calculations performs better on the generality and stability than reducing parameters. However, designing the equivalent alternatives is a very difficult task, which mostly focuses on the approximation of their calculations. Another mainstream solution is compression training. We focus on quantization and sparse methods that have proven effective in practice and are now widely used in large model training or are fundamental designs. Different modules show different sensitivities on the impact of quantization. Therefore, those modules with indistinct errors could be prioritized when selecting low-precision storage and computation, e.g. activation layers. Sparsity is a good implementation of accelerating training, while it requires rigorous hardware support to improve its practical efficiency. The parallel units of the GPU cannot directly apply the sparse calculations, which is the major flaw in the training process. In terms of model initialization, good initial state ensures the stability and efficiency on training large-scale models. Recent advances also suggest that pretraining has great potential to achieve the better performance. To alleviate the huge comsuption in the pretraining, curriculum learning on the model-level plays a key role. It allows the training process to begin with a shallow models or partial parameters and gradually recover the entire model. This technology is going to have a huge boost in other fields for more efficient training.

In the future, we speculate that important scenarios of accelerating training include:
\begin{itemize}
    \item Efficient alternatives to the sophisticated modules. Large biases are easily caused by the accumulated errors of the accelerating alternatives with several individual layers. Considering some sophisticated modules as a whole block to design their alternatives could shrink the output errors.
    \item Efficient operator fusion. The combination of operators in adjacent steps can greatly improve computational efficiency and precision, e.g. for \textit{FlashAttention}~\citep{dao2022flashattention}. These designs make efficient use of the characteristics of the hardware and improve the throughput rate in training, yielding faster speed.
    \item Efficient model initialization. Trainable efficient initialization like pretraining, is a promising research direction in the future. Exploring the interaction of model initialization on different modules is also a worthy direction to study, which will provide the valid guidelines to understand the essence of training the super large-scale models.
    \item Efficient progressive schedule. Designing the novel progressive schedule to guide the training process of large models along all dimensions, could be a promising study in the future. A key issue is exploring the relationship between model size and data samples required for training, which may provide the solid insights for the understanding of the curriculum learning.
\end{itemize}

\section{Optimization-Centric Efficient Training}\label{sec:optimization}


Acceleration schemes for optimization methods have always been a key research direction in the field of machine learning. Reducing the complexity while achieving optimal conditions has always been the target pursued by the academic community. In recent years, efficient and powerful optimization methods have made important breakthroughs in training deep neural networks.\\\\
As the basic optimizer widely used in machine learning, \textit{SGD}-type optimizers successfully help the deep models to achieve various practical 
applications. However, as the problem grows more complex, it is always more likely to fall into local minima and fail to generalize stably. To tackle the difficulties, \textit{Adam} and its variants are proposed to introduce the adaptivity on the updates. This practice has achieved good results in large-scale network training, e.g. for the \textit{BERT}, Transformer, and ViT models. In addition to the own performance of the designed optimizer, it is also important for the combination of accelerated training techniques. We summarize the current thinking on accelerated training based on the perspective of optimization as the following aspects:
\begin{itemize}
    \item \textbf{Learning rate.} Learning rate is an important hyperparameter of non-convex optimization, and it is also critical in the training of current deep networks. Adaptive methods like \textit{Adam} and its variants, have successfully made excellent progress on the deep models. Some strategies for adjusting learning rates based on higher-order gradients also achieve accelerated training efficiently. Implementation of learning rate decay also affects the performance in the training process.
    \item \textbf{Large batchsize.} Adopting large batchsize will efficiently improve the training efficiency. It directly reduces the iterations required to finish one epoch training. At the same time, processing a large batch spends fewer costs less than processing several small batches with the same total amount of samples, for its extra memory utilization and communication bottleneck. 
    \item \textbf{Efficient objective.} Vanilla ERM plays a key role in the minimization problems which enables many tasks to be employed practically. With the deep research of the large networks, several works pay more attention to the gaps between optimization and generalization and propose efficient objectives to reduce test errors. They interpret the importance of generalization from different perspectives and optimize it jointly in the training, which accelerates the test accuracy significantly.
    \item \textbf{Averaged weights.} Weight average is a practical technique to enhance the generality of the models. It considers the weighted average of the historical states with a set of frozen or learnable coefficients, which significantly accelerates the training process.
\end{itemize}

\subsection{Learning Rate}
The learning rate is one of the most important hyperparameters which controls the updates in training. It should be elaborately selected by jointly considering the optimizer and batchsize. From the perspective of the implementation of the adaptation policy, we distinguish them into three types, scaled coefficient, element-wised adaptivity, and matrix-based preconditioner. They adopt a \textit{single value} to decay the learning late, a \textit{vector} to balance the gradients, and a \textit{matrix} to search for optimal respectively.

\begin{table}[t]
    \centering
    \caption{Learning rate adjustment methods.}
    \vspace{0.075cm}
    \renewcommand\arraystretch{1.5}
    \begin{tabular}{|c|c|c|}
    \hline 
    \hline 
    Method & Formulation & Introduction \\
    \hline
    \begin{minipage}[b]{62mm}
        \centering
        \raisebox{-.5\height}{
        \includegraphics[width=50mm, height=28mm]{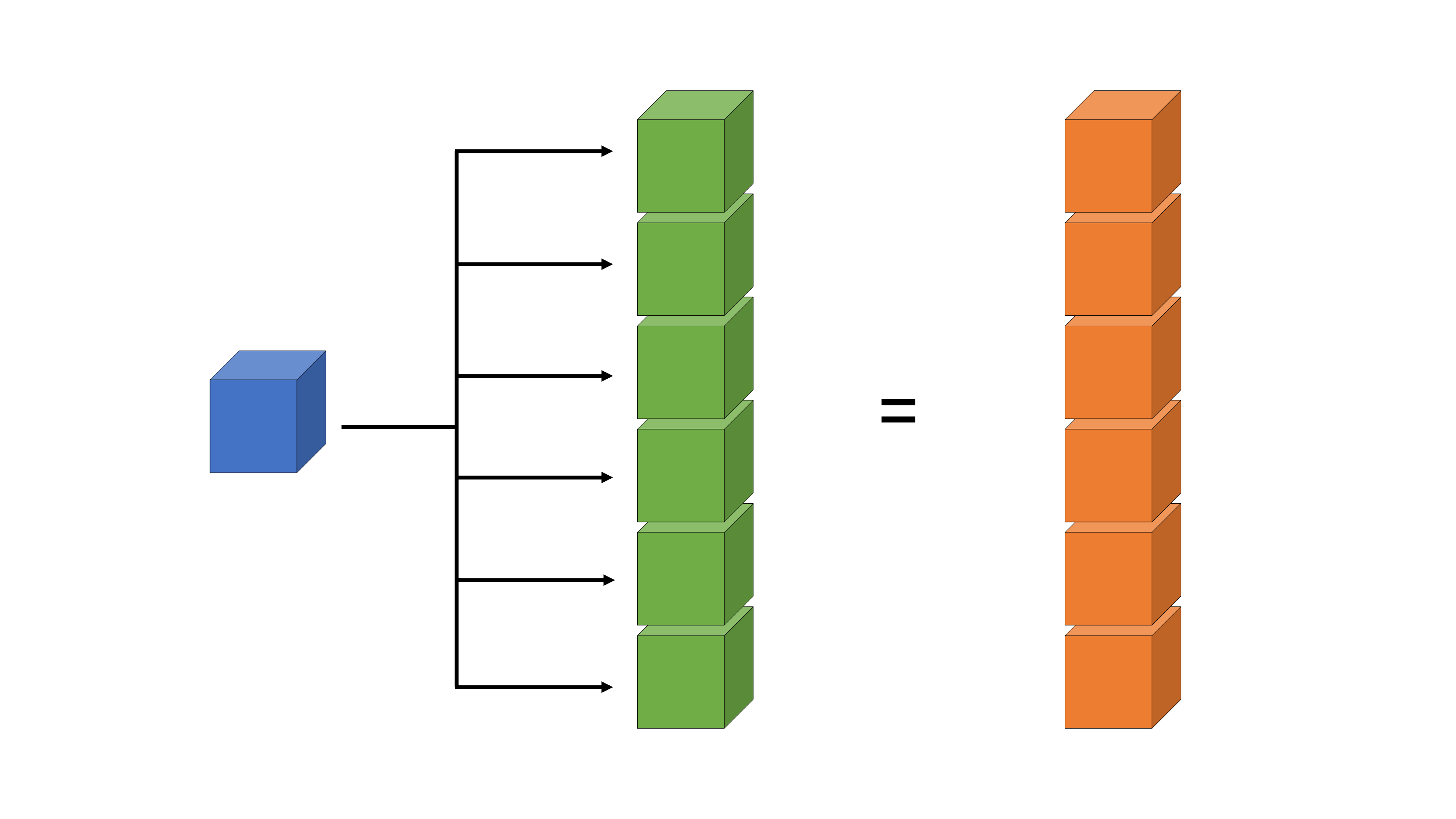}
        }
    \end{minipage}& $\Delta w = \gamma \rho(t) \ g$ & multiplied by a scalar  \\
    \hline
    \begin{minipage}[b]{62mm}
        \centering
        \raisebox{-.5\height}{\includegraphics[width=50mm, height=28mm]{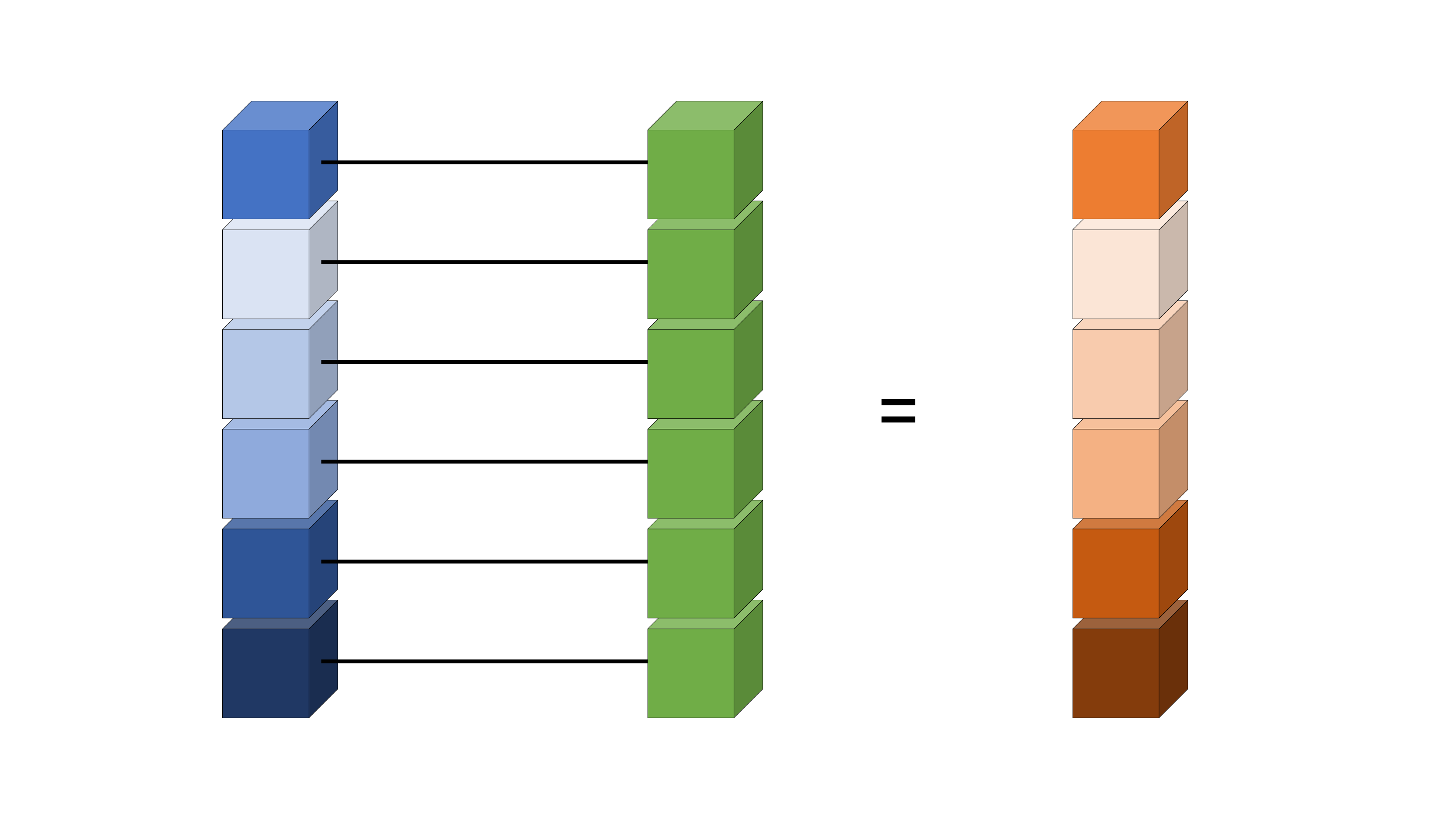}}
    \end{minipage} & $\Delta w = \gamma v(w)\odot g$ & multiplied  by a vector  \\
    \hline
    \begin{minipage}[b]{62mm}
        \centering
        \raisebox{-.5\height}{\includegraphics[width=50mm, height=28mm]{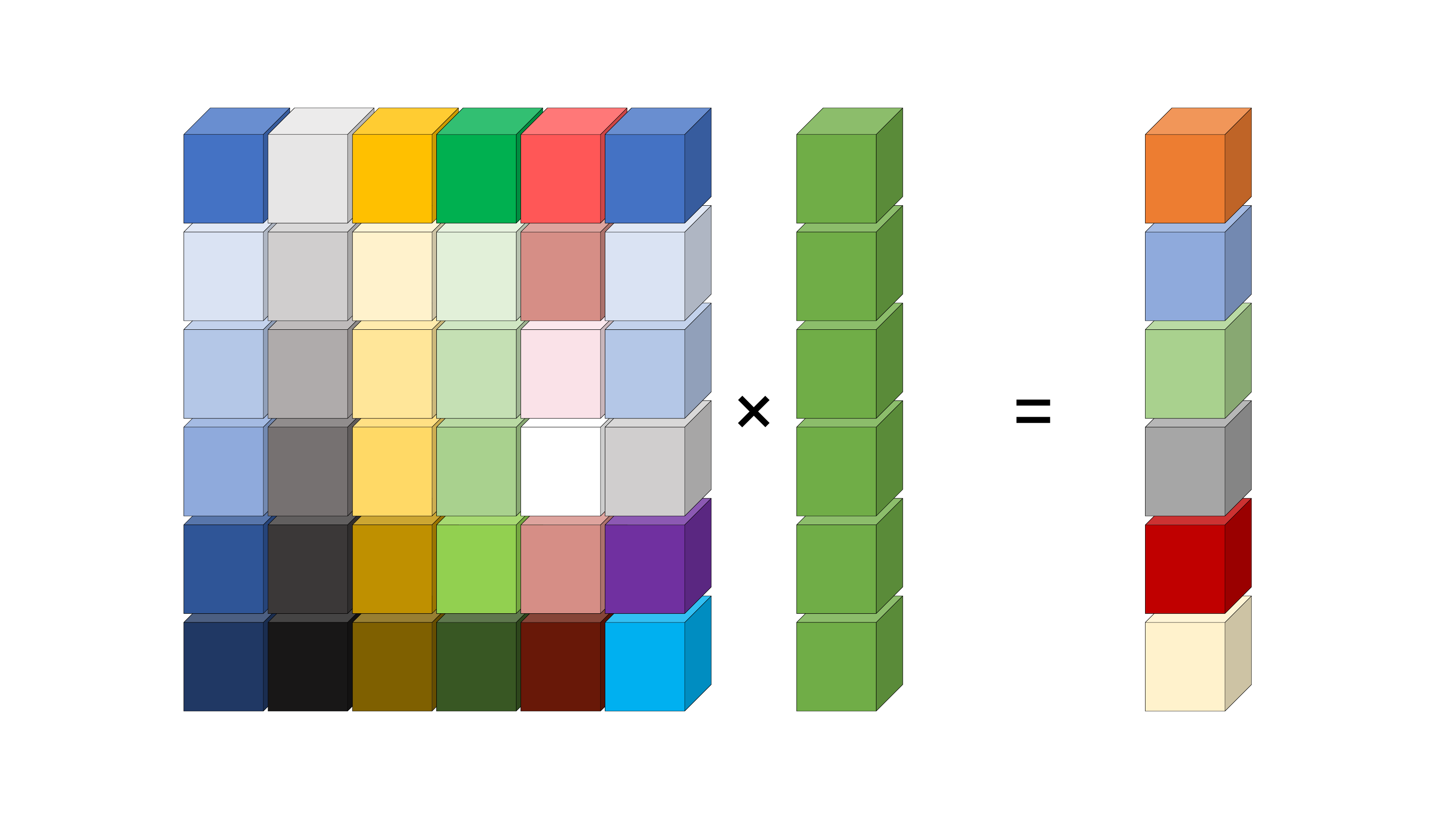}}
    \end{minipage}&  $\Delta w = \gamma P(w)\cdot g$ & multiplied by a matrix  \\
    \hline
    \end{tabular}
    \label{tab:lr}
\end{table}

\subsubsection{Learning Rate Scheduler}
General learning rate scheduler adopts a single decaying coefficient, which gradually reduces the learning rate to learn the fine-grained knowledge of the training dataset. Usually we use epoch as a unit to adjust the change in the learning rate and we can write it as following:
\begin{equation}
    \gamma = \gamma_0\cdot \rho(t),
\end{equation}
where $\gamma_0$ is the initial learning rate, and $\rho(t)$ is a scalar function.

The common scheme is staged-wise decaying with a constant bound, where $\rho(t)=\rho_0^{\lfloor t/t_s\rfloor}$. $t_s$ is the pre-defined decaying stage, which works well in many tasks. However, it fails to outperform on the large models. The simple process of finetuning makes the whole training lack the ability to capture the details, which leads to a mediocre local minimum. 
\citet{loshchilov2016sgdr} demonstrates the negative impact of the simple exponential stage-decayed learning rate. The rapid decrease in the learning rate at the mid-stage of training results in an eventual inability to escape from the local minimum. They propose a cosine decay that the function $\rho(t)=0.5\left(\gamma_{\max}-\gamma_{\min}\right)\left(1+\cos{\left(t\pi/t_s\right)}\right)$, where $t_s$ is the pre-defined schedule to restart the learning rate. The program plays a key role in the training of the current transformer models.
To further improve its performance, a \textit{CLR}~(Cyclical Learning Rates) \citep{smith2017cyclical} policy is proposed to alternately update within the maximum bound and minimum bound. It greatly enhances the generalization performance through periodic adjustments. Different from the \textit{cosine} learning rate, \textit{CLR} adopts a periodic mode with a fixed window. It dramatically accelerates the training speed on the large dataset by reducing the required epochs.
\citet{smith2019super} study the super convergence by expanding the quadratic approximation of the objective loss. A hessian approximation helps to construct its learning rate.
\citet{pan2021eigencurve,extreme-bert} propose a modified \textit{ESD}~(Elastic Step Decay) learning rate scheduler to improve the training for Berts. It adopts a relaxed selection of the learning rate and drives centralized training. Tensile steps provide a more generalized model in the training and create a continuous process of stabilized learning to avoid the local minimum.
\citet{koccyiugit2023accelerating} adopt a scaled 1-cycle learning rate to match the training process with a refined momentum coefficient $\beta_t=\beta_t+0.5\left(\beta_h-\beta_t\right)\left(\cos(t\pi/L)+1\right)$, where $\beta_h$ is the maximum bound of the $beta$ and $L$ is a pre-defined constant to scale it. This combination works better in the momentum-based optimizers and could efficiently improve the test performance.
\citet{ni2022deep} propose a deep model assembling technique to accelerate the training of the foundation model, which separately trains each part of the global model and the merged global model can be fast fine-tuned with the locally trained weights. They show that both Deit-L and Deit-H can be trained faster.

\subsubsection{Adaptivity}
Adaptive learning rate focuses on the element-wised adaptation for all dimensions of the gradient. It considers that every single element of the gradient is related to the objective with a different importance and should be regarded as a local scalar. It contributes to a more accurate and precise controller of the learning rate and works on several large tasks. A variety of adaptivity-based optimizers are designed to improve its efficiency from different perspectives. We will briefly introduce the related works and some novel ideas in the recent advances. The general calculation could be written as:
\begin{equation}
    \gamma = \gamma_0\odot v(t),
\end{equation}
where $v(t)$ is a vector with same dimension as gradients.\\\\
Traditional adaptive methods includes \textit{AdaGrad} \citep{duchi2011adaptive}, \textit{Adadelta} \citep{zeiler2012adadelta}, \textit{RMSprop} \citep{tieleman2012lecture}, and  \textit{Adam} \citep{kingma2014adam}, etc. They utilize a $v(t)=1/\sqrt{\beta v(t-1)+(1-\beta)g_i^2}$ -type function to refine the learning rate. The $v(t)$ function considers that if a single parameter has been updated largely in the past, it should be updated less than other parameters in the current stage, which contributes to balancing the local gradients not to dramatically changed. It helps to stabilize the training process and search for the optimal learning rate according to the historical gradients. \textit{Adam} incorporates the momentum term and adaptive learning rate to achieve higher efficiency and is widely used in many tasks.\\\\
In recent advances, \citet{loshchilov2017decoupled} propose the \textit{AdamW} which considers both the gradient and the weight decay regularization as the momentum term. It has been a great success in transformer training. 
\citet{you2018imagenet} uses the layer-wise regularization \textit{LARS} to facilitate a more accurate scalar on every single layer. Due to the chain rules of the backpropagation, gradients have obvious stratification characteristics, which causes the instabilities from the imbalance between the gradient norm and the weight norm of certain layers. Therefore, it jointly considers the weight norm into the scalar to further finetune the learning rate.
Similarly, \citet{you2017large} use this technique on the \textit{Adam} as the base optimizer and achieve higher performance. To further normalize the $l_2$ regularization term, \textit{AdamW} is proposed to reduce the impact of differences across the parameters of each layer. It approaches great success in the training of the transformer and ViT models.
\citet{adan} propose the \textit{Adan}, the nesterov momentum to accelerate the vanilla \textit{Adam} optimizer, and implement a faster training. It transfers the vanilla gradients into the inertial-type of the gradient to approximate the direction of the next state. Forward-looking further regularizes the gradients, enabling updates to effectively implement accelerated training at an early stage.
 Recently, \citet{chen2023symbolic} propose the \textit{Lion}~(evolved sign momentum) optimizer, which highly accelerates the training speed. \textit{Lion} comes from a policy searching within several pre-defined mathematical function pools. It firstly searches for the sophisticated version of $21$ functions. With the empirical processing and approximating, \textit{Lion} eventually forms the update using a symbolic gradient with momentum. This implementation may be an important research direction in the future.


Adaptive optimizer makes significant progress in the training of large models. Several previous works are based on adopting the element-wise scalars via the Hadamard product. It effectively stabilizes the training process with large batchsize and usually converges faster than the vanilla gradient-based methods. In the future, jointly considering the stage-wise and layer-wise with frozen techniques provides novel insights into the design of the optimizers.

\subsubsection{Precondition}
Preconditioner finetunes the learning rate by a matrix. Beyond the dominant of the first-order gradient methods, the preconditioner involves the high-order derivatives, mainly including the hessian and Fisher Matrix~(FIM). The classical Newton's method applies the hessian to achieve the optimal convergence which accelerates the training efficiently. At present, while such methods are not mainstream in large model training, some of these techniques and insights are worth taking advantage of.\\\\
Simply to minimize the objective $F$, we consider the following adaptation on the learning rate:
\begin{equation}
    w=w-\gamma P(w)\nabla F(w),
\end{equation}
where $P(w)$ is the preconditioner matrix at $w$.

Classical Newton's method adopts the inverse of the hessian matrix as $P(w)=H^{-1}$. However, due to the enormous computational complexity, it is hard to directly benefit from the curvatures. Therefore, the quasi-Hessian methods are proposed~\citep{dennis1977quasi,berahas2022quasi}. They construct the update of the hessian with a Newton function to iterate the quasi-hessian matrix, which reduces the most complexity. To implement the efficient preconditioner, researchers pay more attention on its alternative FIM. It could be proved FIM is the expectation of the hessian on the objective of log-likelihood, which indicates the curvature information to guide the updates. Moreover, it only requires the same computational as the gradients as $P(w)=\nabla l\cdot\nabla l^{\top}$ where $l$ is the log-likelihood of the output distribution. It helps to accelerate the training in several practical scenarios.
\citet{gupta2018shampoo} propose the \textit{Shampoo} method to approximate the FIM tensor as the multiplication of the 3-dimensional matrix. It uses the Kronecker products of two FIM information, which could be calculated in block-wise structures. Compared with adaptivity, it has more tension in depicting the characteristics at the state $w$. Similar work like \textit{K-FAC}~\citep{martens2015optimizing} which proposes to employ the multiplication of the expectation of the inputs and gradients respectively instead of the expectation of their multiplication. Another impactful attention focuses on matrix-free approximations. 
\citet{frantar2021m} study the approximation by its definition which records the adjacent difference of the gradients and calculates their product. The preconditioner matrix is estimated as a sum of several rank-one matrices via a pre-computation process and inverse alignment, with lower computational overhead than the non-matrix-free methods.
\citet{pauloski2021kaisa} present the \textit{KAISA} framework to implement the memory footprint and hardware support on the communication and computation efficiency. It rebuilds the process of computing FIM and the infrequent updates on the curvature information to further stabilize the training.

It is a subject of concern to study the preconditioners which could be considered as a matrix projection of the learning rate to achieve the optimal convergence. Due to the complexity of the calculations, it is not yet working efficiently on large models. Designing a preconditioner matrix with first-order computational complexity to accelerate the training of large models is an important target in the future. And, the objective of balancing the optimization and generalization may provide more novel insights.

\subsection{Large Batchsize}
Appropriately adopting large-batch training could reduce time consumption, improve the utilization of storage, generate smooth gradients, and maintain a stable training process. However, training with excessive batchsize is counterproductive which may dramatically damage the stability of the training and even leads to divergence, especially on large-scale models. 
\citet{keskar2016large} show that small batch training tends to converge to flat minima farther from the initial state, and large batch training tends to converge to the sharp minimal around the initial state and generalizes worse on the test dataset. Therefore, it is an important study to learn how to train with large batchsize while ensuring performance.
\\\\
To discuss the batch size term in our equation, we will focus on mainly how to enable large batch-size training as it is a strategy applied by most of the state-of-the-art models. Let us start with a brief introduction to Stochastic Gradient Descent (SGD) \citep{robbins1951stochastic}, a commonly used and important method for deep learning. 
The iterative step of minibatch gradient descent is shown as follows:
\begin{equation}
    w_{t+1}=w_{t}-\frac{\eta_t}{B_{t}}\sum_{i=1}^{B_t}\nabla{f_{i}(w_{t};(X_i,Y_i))},
\end{equation}
Where $B_{t}\subset B$ is a minibatch sampled (generally uniform sampling) at the $t$-th step from an entire training dataset $B$. Acceleration can be achieved by overcoming two aspects of shortcomings of SGD, the inefficient sampling, and the challenging parallelism.\\\\
To achieve training with large batches and avoid poor generalization performance, some existing solutions focus on jointly finetuning the learning rate and batchsize, including linear scaling of the learning rate as the batch size expands, adopting the warm-up of the learning rate at the beginning stage, and gradually increasing the training batchsize. Moreover, extra augmentation on the data and the proposed efficient optimizers could also support the large batch training, which enhances the generalization performance and accelerates the training process. 
\citet{smith2017don} show that increasing the batch size naturally leads to similar results as the expeditiously decayed learning rate, which could reduce the unexpected errors in the parameter updates, facilitate its parallelism, and achieve higher training efficiency.
\citet{hoffer2017train} show using specific normalization to control the covariance shift in training could help to stabilize the training and reduce the gap between generalization and optimization.
\citet{mccandlish2018empirical} propose an empirical statistic called the gradient noise scale to predict the largest batch size for the model. Consider a model parameterized by $w \in \mathbb{R}^d$, let $G$ denote the true gradient, $H$ denote the true Hessian at $w$, and $\Sigma$ denote the per-example covariance matrix, the gradient noise scale $\mathcal{B}_{\text{noise}}=\frac{\text{tr}(H\Sigma)}{G^\top H G}$, simplified as $\mathcal{B}_\text{simple}=\frac{\text{tr}(\Sigma)}{{\vert G\vert}^2}$ if under the assumption that the optimization is perfectly well-conditioned (Hessian is a multiple of the identity matrix), and it predicts the critical batch size $\mathcal{B}_{\text{crit}}$. The gradient noise scale is defined as scalar to balance the trade-off.

For efficient distributed systems, a similar idea to gradient accumulation is used to reduce training time by dividing the computation of gradients among multiple machines in parallel, and some problems such as learning rate, optimizer, and communication need to be solved, which leads to efficient parallelism in the cross-device training. \citet{goyal2017accurate} train ResNet-50 with 256 GPUs and increase the minibatch size from 256 to 8192 with the same accuracy in only 1 hour of training time (the original work took 29 hours to train with 8 Tesla P100 GPUs). They also use a gradual learning rate warm-up and propose random shuffling using a single distribution per epoch partitioned by the number of workers.
\citet{you2018imagenet} use the layer-wise scaling~\citep{you2017large} to increase the batch size to 32K and utilize 1024 CPUs for ImageNet training. They train the AlexNet for 100 epochs in 11 minutes to achieve comparable performance.
\citet{jia2018highly} show a training system including the properties of (a) a mixed-precision training method that improves the single GPU training throughput without losing accuracy, (b) an optimization method for extremely large minibatch size, and (c) highly optimized all-reduce algorithms that speed up the backward process compared to NCCL-based training.
\citet{li2022stability} study the stability-efficiency dilemma which is difficult to overcome with the current techniques. They propose a sequence-length warm-up strategy to train the GPT models with a larger batchsize to avoid these negative impacts.
\citet{xue2022large} indicate the serious challenges on the complicated pipelines of dense visual predictions via utilizing the large batch training and propose an efficient adaptive gradient variance modulator which could enable the stable generalization performance with a very large batchsize~(i.e. $10K$). 

Large batch training is a widely used acceleration technique, but it requires selecting fine-grained hyperparameters like learning rate to prevent falling into the unexpected local minimal~\citep{he2021large}.
As mentioned in~\citet{golmant2018computational}, increasing the batchsize after a specific state (critical batch) leads to computational inefficiency and fails to speed up convergence. Meanwhile, if the aggregated batchsize across the devices participating in training is too small, the additional waiting conflicts occur during the communication~\citep{smith2022using}. Therefore, as the training data and size continue to increase, to further implement efficient training on large-scale models, the adaption of large batchsize is still a hot topic worth exploring further.

\subsection{Efficient Objective}
The objective is always one of the most important factors in the optimization, which is a principle we want to optimize for the tasks. Optimization objectives directly determine the ultimate targets and training efficiency. In this part, we review the optimization-centric acceleration techniques which benefit from the considerations on designing novel objectives. Instead of the narrowly defined objectives, we summarize the methods which aim to construct the update vector $G$ in equation~(\ref{eq:gradient-based descent}) to provide a comprehensive review.\\\\
Empirical risk minimization (ERM) is a principle in statistical learning theory that defines a family of learning algorithms and is used to provide the theoretical bounds on their performance. In solving minimized machine learning tasks, various methods for optimization have been developed significantly. However, it does not prevail in balancing the optimization efficiency and generalization error. Because the optimization objective and the expected target in practice are biased, the ERM is with low tension and lacks precision on the representation of generality. Therefore, a series of amended objectives are proposed as the global target jointly considering both the efficiency and stability~\citep{scieur2016regularized}.
In addition to the widely used optimizers, such as SGD, Adam, etc., layer-wise adaptive rate scaling~(\textit{LARS}) is also a general approach aimed at training with large minibatch. It allows different modules or layers to adopt a self-adaptive learning rate instead of a fixed one. Fine-grained adjustment efficiently alleviates the local over-fitting with the large minibatch.
\citet{khosla2020supervised} adopt a \textit{LARS} selection on the supervised contrastive learning on different modules to stabilize the training process and it achieves a better performance on the modern batch contrastive approaches compared to the traditional contrastive losses such as triplet, max-margin, and the NN-pairs loss. A similar idea can be found in~\citep{chen2020simple}. 
\citet{goyal2021self} try a self-supervised pretraining mechanism on the uncurated ImageNet dataset. The \textit{LARS} technique makes the huge models surpass the best self-supervised pre-trained models. It helps to deal with the inputs in different resolutions and avoid the parameter collapse that is easy to occur in the early stage of training.
\citet{foret2020sharpness} propose \textit{SAM}~(Sharpness-Aware Minimization), which simultaneously minimizes loss value and loss sharpness to improve generalization. It reveals that vanilla ERM leads to a sharp minimal with low generalization performance. Therefore, \textit{SAM} considers optimizing the worst point within the ball at the current state in each iteration. Jointly considering this with its sharpness, it searches for the flat landscape of the optimized state. \citet{andriushchenko2022towards} propose an understanding of SAM and propose a variant that updates with both the SGD and SAM gradient. One expected improvement of the SAM optimizer is how to reduce the twice gradient calculations which may cause extra costs in the training.
\citet{mindermann2022prioritized} introduce the \textit{RHO-LOSS}~(reducible holdout loss selection) to accelerate the training on the web-scale data. They focus on the optimization performance of the holdout dataset which shows the same distribution with total samples. To reuse the cross-entropy loss on the vanilla training samples, it could be written as the final formation of the loss difference between the training set and the holdout set. The second part is approximated as the training model without the training distribution which can efficiently save the calculations. Therefore, it enlarges the batchsize and selects those bad points to train in a single iteration, which could speed up the training process.

Objective improvements consider more insights and practical issues in the vanilla ERM. How to represent the generalization gaps and errors is becoming a research topic worth studying in the future. Combined with empirical evaluation methods, using large-scale search to find the optimal optimizer shows significant efficiency, within a reasonable searching space.



\subsection{Averaged Weights}

Model averaging is an important technique to generate a generalization-efficient model via several historical states. The generated models usually show higher test accuracy than the training models. The average model does not require training and is used only for testing, which introduces no extra costs on the calculations and memory utilization.\\\\
Model average comes from the \textit{Anderson acceleration}~\citep{anderson1965iterative, walker2011anderson} and makes good progress on the stochastic non-convex optimization.
\citet{scieur2017nonlinear} propose the \textit{RNA}~(Regularized Nonlinear Acceleration) to aggregate the historical states via a set of optimized weights which are solved by a minimization on a regularization term. It searches for the scaled coefficient for every single state and merges them as the next initial point. 
\textit{EMA}~(Exponential Moving Average) is a simple model merging technique that uses an exponentially decaying weight to calculate the weighted average of the parameters at each previous training step. It significantly improves the test accuracy in the middle stage of the training. 
\citet{izmailov2018averaging} propose the \textit{SWA}~(Stochastic Weight Averaging), a model averaging technique that captures model weights according to a constant or cyclical learning rate and maintains an average of the weights to reduce the generalization error. Compared with the \textit{EMA}, \textit{SWA} remembers all the historical states and averages them, which provides a flat minimal. They show that for Top-1 Accuracy with ResNet-50, ResNet-152, and DenseNet-161 on ImageNet, it provides about 0.6-0.9$\%$ improvement over the pre-trained models.
\citet{li2022trainable} propose the \textit{TWA}~(Trainable Weight Averaging) to further improve the generalization performance, which is based on the projection of a set of orthogonal bases. It works in a reduced subspace spanned by several historical states which are usually frozen at the beginning of the training process. It largely reduces the approximation error in \textit{SWA}. It is easy to implement and could be flexibly plugged into different parts of training.
\citet{kaddour2022stop} study a truncated moving window with a fixed size to average the historical states, which is named \textit{LAWA}~(Latest Weight Averaging). It indicates that only the last few states help to generalize well. The selection of the states could not be the adjacent points one by one, and it selects each state after a total epoch training. This model does not play a key role in the early stage of training of the large variance of the models. When the model stabilizes, the performance improvement of \textit{LAWA} is excellent with usually higher acceleration.

Model average helps to approach a set of better-generalized states during the training process. Although such methods are not directly involved in optimizing updates, their significant improvements in the generality help to early stop the training with the comparable parameters. It dramatically reduces the required training epochs. In the field of efficient training, it is a promising study in practical applications.

\subsection{Discussion}
In this section, we review the general acceleration techniques from the optimization-centric perspective. We mainly focus on the practical applications of selecting an efficient learning rate to accelerate training, adopting large batch training, designing efficient objectives, and utilizing the model average techniques for better generality. These techniques imply two mainstream scenarios of accelerating the optimization process and increasing the generalization performance. We provide a comprehensive analysis of each component.

Recent advances in the selection of the learning rate indicate that adaptivity is the right hand to maintain high efficiency with acceptable extra computational costs, which benefits from the element-wise scalars on the different dimensions of gradients. One major problem to note is that scalars of squared momenta may cause a sharp diminishing of the learning rate, yielding much slower convergence. The application of a preconditioner is a promising approach to introducing the correction of the curvature which may lead to fast convergence. However, it suffers from the expensive costs to update the preconditioner in practical training, especially on large-scale models. Meanwhile, inaccurate curvature direction may mislead the optimization into the local minimum. Large batch training shows a strong dependence on the selection of hyperparameters. The solution to efficient large batch training usually adopts adjusting other hyperparameters or designing novel objectives. These two routes focus on the acceleration of the optimization process, which encourages achieving a better convergence.

The other scenario pays more attention to improving the generalization efficiency. Many previous works have studied the gaps between optimization and generalization, and propose to search for the appropriate state with a flat loss surface. Several objectives based on the valid measurement strategies for flatness are proposed to enhance the generality of the large-scale models and achieve great success. Although some additional computational consumption cannot be overlooked, this objective-based improvement is still very effective. Compared with objective design, model averaging is a straightforward and effective method to improve generality. It can achieve effective acceleration without introducing any additional computational costs and shows great potential in the middle stage of training. As the training is completed, its improvement eventually disappears. 

In summary, we perceive that future developments from the optimization-centric perspective include the following directions:
\begin{itemize}
    \item Refined adaptive learning rate. Further balancing the biases caused by erratic gradients when training large-scale models help to achieve higher efficiency. Considering curvature information to construct low-cost preconditioner matrices for fine-grained adjustments is worth exploring, which may facilitate the adaption of larger batchsize in the training.
    \item Valid measurements on the generality of the training models. Several measurements are adopted to evaluate the local stability and flatness, e.g. top eigenvalue of hessian, gradient norm, the worst state in the neighborhood, and achieve great efficiency in the training accelerations. Exploring low-cost representations helps to refine the training process with high generalization performance. Meanwhile, utilizing reasonable awards and policies to search for the best combination of the updated formulation shows great potential for improving training efficiency~\citep{chen2023symbolic}. 
    \item Training involvement of averaged models. The averaged model with good generalization performance can be considered a better initial state for the next iteration. Designing better optimization algorithms to make full use of this aggregation state will be a promising direction to follow.
\end{itemize}

\section{Budgeted Efficient Training\label{sec:budgeted_training}}

Recently, several works focus on training deep learning models with fewer resources but achieving higher accuracy as much as possible. 
This type of problem is defined as budgeted training, i.e., training under the given budget (a limitation on the measurable costs) to achieve the highest model performance~\citep{li2019budgeted,pan2021eigencurve}.
To systematically consider the hardware support to get closer to the real situation, we define budgeted training as training with given devices and limited time, e.g., one day on a single low-end deep learning server~\citep{izsak2021train,geiping2022cramming}, to obtain the model with the best performance. 
The study of budgeted training can shed light on how to produce training recipes for budgeted training, including deciding on the configuration of model size, model structure, learning rate schedule, and several other tunable factors that affect performance, as well as combining efficient training techniques that fit the available budget.
In this section, we mainly review the advanced developments in the budgeted training.

\subsection{Techniques for Budgeted Training}
Due to impacts introduced by the budgets, researchers have developed many more targeted technologies for efficient budgeted training.
In this part, we summarize the approach to budgeted efficient training from the perspectives of data, model, and optimization. 
Then, we introduce the process of providing a budgeted training recipe and summarize techniques proposed or adapted for budgeted training.

\paragraph{Data.} 
From the data perspective, budgeted training focus on the budget allocation regarding the dataset and the algorithms related to the data.
Scaling up either the amount of data processed in training or the model size can improve training performance, as \citet{kaplan2020scaling} show that there is a power-law relationship between model performance and each of the following three factors: the number of non-embedding model parameters $N$, the training dataset size in tokens $D$, and the amount of non-embedding compute $C$.
However, with the same amount of computations $C$, there is a trade-off between $N$ and $D$, e.g., a larger model processes fewer data per unit of compute.
\citet{kaplan2020scaling} propose that there exists an optimal budget allocation between the model size and the amount of data processed in training. They demonstrate that within a fixed compute budget (the term ``compute budget'' refers to the total amount of computations), the optimal model performance is obtained by training very large models and stopping early before convergence.
\citet{hoffmann2022training} show different results and propose that given the compute budget, the number of data tokens processed in training should be scaled equally to the size of the model, instead of only scaling models but keeping the amount of training data constant, as in previous work.
They show that smaller models that are adequately trained can overperform undertrained large models, and they test this hypothesis by training the language model Chinchilla (70B parameters, $4\times$ more data), which achieves a better score than another large language model Gopher (280B parameters) with the same compute budget. 
The above work summarizes the empirical laws for deciding the size of the dataset under a fixed budget. 
Furthermore, \citet{steiner2021train} study the interactions between the amount of training data, \textit{AugReg} (data augmentation and model regularization), model size, and compute budget. 
They show that the model trained with \textit{AugReg} and a smaller dataset but with more compute budget reaches the same performance as the model trained on more data. 
For limited data scenarios, \citet{chen2021data} propose a data-efficient GAN training method that identifies trainable sparse subnetworks ("winning tickets"~\citep{frankle2018lottery}) from the original GAN using the small training set of real images. This subnetwork is then further trained using the same limited dataset with data- and feature-level augmentation techniques. They show that their framework complements real image data augmentation methods.
In addition, \citet{zhao2020differentiable} propose Differentiable Augmentation (DiffAugment) for GAN training, which applies differentiable augmentation techniques to both real and generated samples. They show that this improves stability and convergence performance, and is particularly effective when training data is limited.
It should be noted that several techniques used for data processing are computationally intensive, such as data augmentation and importance sampling, leading to trade-offs when using these techniques for a given computational budget. 
\citet{arazo2021important} study the interaction of importance sampling and data augmentation in budgeted training of a fixed number of iterations, showing that adequate data augmentation leads to better performance than importance sampling methods and leads to speedups. 

\paragraph{Model.} 
As we mentioned, given a compute budget, the optimal number of model parameters and the amount of data can be derived, and \citet{kaplan2020scaling} suggest training very large models on moderate data and stopping early before convergence. 
\citet{izsak2021train,geiping2022cramming} also demonstrated that training large models rather than small ones is suitable for low-budget situations.
Similar findings are supported by \citet{li2020train}, who propose a training strategy that trains a larger model (wider and deeper) and stops early, then compresses it heavily with quantization and pruning, instead of training a smaller model until convergence with the same budget. 
They also show that larger transformer models converge much faster than smaller transformers, and are more robust to compression. 
Moreover, there are other benefits of training large models, such as \citet{wei2022emergent} show that scaling up can bring additional (emergent) capabilities for language models. 
Having decided on the number of parameters for the model, it is still possible to find and optimize effective architectural changes to the model to accelerate budgeted training or improve performance.
\citet{izsak2021train} investigate the pretraining of a BERT-like masked language model in 24 hours using a single low-end deep learning server. They present a training recipe including using training large models, using sparse token prediction (predicting only masked tokens)~\citep{liu2019roberta}.
With the budget of training on an RTX 2080Ti for 24 hours, which is similar to the budget set by~\citet{izsak2021train}, \citet{geiping2022cramming} demonstrate that removing biases from the QKV projections of the attention block and disabling all linear layer biases improve the computation efficiency. 
They also examine that applying layer normalization before the attention and FFN blocks~\citep{baevski2018adaptive}, using Gated Linear Units~\citep{shazeer2020glu} can improve performance. 
Recently, \citet{panbudgeted} observe from the perspective of the model structure that the Transformer model exhibits different levels of redundancies at different training stages. They propose a budgeted training framework that adjusts the duration on each level of model complexity by controlling the activation rate of the model structure, mainly the number of attention heads, the hidden dimension of the MLP, and the number of visual tokens during training, under a fixed budget. 

\paragraph{Optimization.} 
\citet{izsak2021train} show several important hyperparameters to optimize for higher performance in budgeted training, including batch size, peak learning rate, warmup proportion of the learning rate, and the number of days for training. 
The hyperparameters in budgeted training are configured differently than in the large-scale framework and need to be adjusted to fit a fixed budget.
To find the best hyperparameter configuration, it may be necessary to perform several tests with different hyperparameters within the estimated hyperparameter range. 
In addition, several studies focus on efficient training schedules.
\citet{li2019budgeted} propose the budgeted training setting and show that a new learning scheduler (vanilla linear scheduler) could be adopted to improve the performance, particularly for small budgets. 
\citet{chen2022rex} propose the combination of a new function of the learning rate schedule and a sampling rate to update the learning rate by sampling from the given function. They show that their schedule outperforms the linear schedule in low-budget settings.


\paragraph{Overall recipe for budgeted training.}
The overall recipe for budgeted training should include configuring key hyperparameters and establishing a training pipeline that incorporates several efficient training techniques. 
To initiate budgeted training, it is important to decide on several hyperparameters that significantly affect the final performance. The total computational budget can be estimated based on the limited training time and the computational capacity of the device being used. Following this, the model size and the number of training data tokens can be determined based on empirical laws~\citep{kaplan2020scaling,hoffmann2022training} or existing recipes. 
To determine the optimal configuration of parameters, experiments can be conducted to observe early training results and select the best set of parameters.
At the data level, after determining the size of the dataset, one can collect an appropriate amount of high-quality data and apply necessary processing, such as data augmentation techniques.
At the model level, although the scaling laws~\citep{kaplan2020scaling} suggest that changing the model structure only yields limited performance improvements once the model size is fixed, computational efficiency can still be enhanced with certain model structure changes~\citep{geiping2022cramming}.
At the optimizer level, there are also some hyperparameters, such as the peak learning rate and warmup proportion of the learning rate schedule, which should be decided based on the existing formulation or by hyperparameter search. And the efficient optimizers should be introduced to schedule learning rates~\citep{chen2022rex} or model structures~\citep{panbudgeted}.
Finally, from a systematic perspective, efficient techniques suitable for the given budget can be merged into an existing recipe. Techniques optimized for the given hardware can be used to improve system performance, such as sharding dataset offline and loading into RAM~\citep{izsak2021train} to reduce disk I/O bottlenecks and using disk serialization~\citep{extreme-bert} to save RAM. These techniques may help to improve the throughput rate of the system and achieve an overall speedup without loss of accuracy.
In addition, as we will describe in section~\ref{sec:system}, frameworks such as DeepSpeed~\citep{rasley2020deepspeed} can be added to the model to implement the ZeRO optimizer~\citep{rajbhandari2020zero}, automatic mixed precision training, and fused implementations. 
Taken together, these techniques can provide an efficient training recipe.
Excluding the configuration of hyperparameters, we summarize and present the table \ref{tab:budgeted} as a budgeted training recipe consisting of various techniques.

\begin{table}[t]
\footnotesize
\centering
\caption{Techniques to improve speed and availability for budgeted training.}
\label{tab:budgeted}
\vskip 0.075cm
\begin{tabular}{llll}
\toprule
Centrics & Aspects  & Methods & Benefits \\
\midrule
Data & dataset & short sequence & budgeted computation and memory\\
         & dataloader & disk serialization~\citep{extreme-bert} & budgeted memory, save RAM \\
         & & sharding offline and loading into RAM~\citep{izsak2021train} & avoid disk I/O \\

\midrule
Model & model architecture & diabling QKV and linear biases~\citep{geiping2022cramming} & computation efficiency\\
            & & \textit{PreNorm}~\citep{shoeybi2019megatron,izsak2021train} & stability, large learning rate \\
            & & Gated Linear Units~\citep{shazeer2020glu,geiping2022cramming} & improving performance \\
            & & removing nonlinear attention head~\citep{geiping2022cramming} & computation and memory efficiency \\
            & & complexity schedule~\citep{panbudgeted} & efficiency \\
            & model size & training large models & efficiency \\

\midrule
Optimizer & learning rate schedule & linearly decaying to zero~\citep{li2019budgeted}  & fast budgeted convergence \\
                    & & one-cycle learning rate~\citep{smith2019super} & fast convergence \\
                    & & Elastic Step Decay~\citep{pan2021eigencurve,extreme-bert} & pretraining acceleration \\
                    & large batch size & gradient accumulation & computation efficiency \\
                    &                       & gradient checkpointing & budgeted memory \\
                    &                       & aggresive batch size schedule~\citep{geiping2022cramming} & progress in early training \\ 
                    &                       & importance sampling & computation efficiency \\
                    & weight averaging & latest weight averaging~\citep{kaddour2022stop} & speedup \\
                    & others & sparse token prediction~\citep{liu2019roberta,izsak2021train,geiping2022cramming} & memory efficiency \\
                    & & gradient clipping & stability \\
                    & & disabling dropout during pretraining~\citep{geiping2022cramming} & computation efficiency \\
                    & & early stopping & efficiency \\


\bottomrule
\end{tabular}
\end{table}

\subsection{Discussion}

In this section, we have reviewed efficient training techniques from the perspective of budgeted training.
We summarize techniques for efficient budgeted training in terms of data, model, and optimization, respectively, and present a comprehensive recipe of techniques.

The idea of trade-offs, i.e., sacrificing one for the gain of the other, plays a significant role in budgeted training. 
For instance, techniques that mitigate memory limitations through space-time trade-offs have been developed.
Moreover, given the limited budget for training, there are often trade-offs between different techniques and hyperparameters that affect model performance.
When these parameters are individually scaled or when efficient techniques are used in isolation, they can usually improve model performance.
But with the fixed compute budget, trade-offs need to be made between these hyperparameters, and between computationally intensive techniques. 
Several systematic studies that examine various training configurations provide empirical recommendations for deciding on the trade-offs in practice and have suggested effective budget allocation schemes~\citep{kaplan2020scaling,hoffmann2022training}. 
The problem is that the conclusions may still be one-sided, i.e. valid only for parameters in a certain range, or the conclusions give biased laws because they are only empirical laws drawn from experiments.
These laws may still require future work to train more models to validate them in new settings, which will incur more costs than following the configuration of existing model training to get good results.
It seems relatively feasible to start with small-scale, low-budgeted training setups to decide on an efficient budgeted training recipe and then scale up to larger scales.
Another idea that may be valuable is to design more budget-aware training components, i.e., training components that can adjust parameters according to the remaining budget, such as the learning rate schedulers proposed in some recent studies~\citep{chen2022rex}.


In summary, we recommend that future research on budgeted training focus on the following directions: 
\begin{itemize}
    \item Practical benchmarks.
    For this, we mean that it is possible to study an identical and practical budget benchmark, e.g., a day of training on a low-end GPU server~\citep{izsak2021train,geiping2022cramming}. 
    It is too expensive for researchers to conduct a systemic study under different configurations (such as \citep{kaplan2020scaling,steiner2021train,hoffmann2022training}). Instead, for the same practical budget, future work can follow the existing training recipe and introduce further optimizations.
    \item Budget-aware schedulers. Algorithms such as efficient learning rate schedules~\citep{chen2022rex} can be tuned to the remaining budget, which facilitates fast convergence and improves performance. More techniques that can dynamically respond to the remaining budget are still to be developed.
\end{itemize}
  
\section{System-Centric Efficient Training}\label{sec:system}
Researches from the system-centric perspective provide the specific implementation methods for the designed algorithms. It studies the valid and practical executions of the hardware which can truly achieve efficient training. In our survey, we focus on the implementation of general computational devices, e.g. CPU and GPU devices in the multi-nodes clusters. Solving potential conflicts in the designed algorithms from the perspective of hardware is the core concern. In this section, we mainly review the techniques for hardware implementation in existing frameworks and third-party libraries, which efficiently support the processing of data, models, and optimization.
Then we introduce some of the existing open-source platforms, which provide solid frameworks for model building, efficient use of data for training, mixed precision training, and distributed training. This section is organized as follows:
\begin{itemize}
    \item \textbf{System-centric Data Efficiency.} Efficient data processing and data parallelism are two important concerns in systematic implementation. With the rapid increase of the dataset, inefficient data processing gradually limits the training efficiency, especially in large-scale training on multi-nodes. Designing more hardware-friendly computing methods and parallelization can effectively avoid time wasting in training.
    \item \textbf{System-centric Model Efficiency.} As the number of model parameters expands dramatically, system efficiency from the model perspective has become one of the important bottlenecks. Storage and computational efficiency of large-scale models bring great challenges to hardware implementations. In this part, we mainly review how to achieve efficient I/O of deployment and streamlined implementation of model parallelism to speed up the practical training.
    \item \textbf{System-centric Optimization Efficiency.} The optimization process represents the backpropagation and update in each iteration, which is the most time-consuming calculation in the training. Therefore, the implementation of system-centric optimization directly determines the efficiency of the training. To clearly interpret the characteristics of system optimization, we focus on the efficiency at different computing stages and review the improvements of each process.
    \item \textbf{Open Source Frameworks.} Efficient open-source frameworks facilitate training as a bridge to grafting algorithm design and hardware support. We investigate a range of open-source frameworks and analyze the strengths and weaknesses of each design. 
\end{itemize}

\subsection{System-centric Data Efficiency}
Although data processing does not have a high proportion of time costs in training, inefficient handling still leads to drastic increases in training time. Especially when the large batch is trained on the large-scale model, rational use of non-conflict processes to accelerate data processing can effectively reduce additional consumption. A lot of work has been proposed at the system level to achieve effective data processing. In this part, we mainly review the efficient implementation of data processing and data parallelism. Then we introduce some efficient libraries which could be adopted in practical scenarios.\\\\
Efficient data processing can bring significant benefits to accelerated training, which achieves high-efficiency training and saves redundant wait time in the system.
\citet{bai2021efficient} propose the \textit{PaGraph} to support general and efficient sampling-based graph neural network training on multi-GPU devices. It exploits available GPU devices to maintain the frequently-accessed graph data with cache storage. 
\citet{qi2023speechain} introduces \textit{SpeeChain}, an open-source PyTorch toolkit to develop the large-scale machine speech chain application. It supports both the multi-dataloader batch generation and on-the-fly data selection techniques.
\citet{jia2022data} design the \textit{A-Dloader} to dynamically allocate the CPU resources to concurrently run multiple programs and adopt the dataloader allocation as the local valve to reduce the visiting intervals of the GPU devices, which improves the overall efficiency.
\citet{svogor2022profiling} evaluate several benchmarks to outline the performance issues within certain steps of the dataloader pipeline and modify the \textit{ConcurrentDataloader} to further improve the GPU utilization. It supports the efficient training of cloud-based datasets.
\citet{xiedeep} focus on the overlap datasets executed by multiple training tasks in parallel. They propose a dependent sampling method \textit{JOADER} supported by the domain-specific cache policy and a novel tree structure for the loading data. 
\citet{dash2022scalable} develop a solution for supercomputers to alleviate the low loading efficiency of the small patches in the very high-resolution images, extreme sparsity of the notations, and I/O latency on the disk storage. 

In the case of multiple GPUs, data parallelism can be used to speed up the computation. With data parallelism, each node has a copy of the model, takes a different shard of the data, and then performs forward and backward computations to obtain the gradient. 
In a centralized architecture, these nodes send their computed gradients to the parameter server, which aggregates the parameters, updates the model, and sends the parameters back to each node. 
In contrast, in a decentralized architecture (more common in the case of deep neural network training), each node updates its corresponding part of the model weight after performing a multi-node communication. 
As the models get larger, distributed training becomes an engineering problem, and the trainer needs to optimize the distribution strategy, as well as network communication and memory.
For example, training PyTorch models on multiple GPUs can be done by simply modifying the code to use \textit{torch.nn.DataParallel}. However, when training large models such as a language model on large batches, it is necessary to balance the GPU load, e.g., by further distributing the loss criterion computation to avoid overloading the server nodes that collect the results.

\paragraph{Efficient Libraries.} Efficient libraries provide a solid support for the training process. When the dataset size is small, all data samples can be stored in the computing device before training, which spends no extra costs on the data I/O. However, with the rapid development of deep neural networks, the size of the dataset grows geometrically. Such a huge amount of data causes them to be stored on the local storage, and the samples are alternately read in batches as they participate in training. In the practical scenarios, dataloader is not the serious bottleneck~(usually lower than $10\%$ of the total).
Traditional dataloader in each framework performs data I/O and preprocessing in the CPU with the multithreading lines, which requires a guaranteed number of cores. 
\citet{leclerc2022ffcv} design the novel structure to store the image sample in the \textit{FFCV} as the \textit{.beton} file with four major sections of the header, data table, heap storage, and allocation table. They recombine and compile each section in the whole process to make full use of the cache structure and mechanism. Even in the case of particularly expensive disks, a quasi-random sampling technique is adopted to minimize the stress on the underlying storage. Another core tenet is to avoid unnecessary memory allocation in the pipeline with a circular buffer. This boost makes it very efficient when processing digital sources like images and videos.
Several data augmentation methods involve various basic matrix calculations, which leads to another novel acceleration in data augmentation. NVIDIA proposes the Data Loading Library~(\textit{DALI}) to support fast, portable, and flexible data loading. \textit{DALI} is an open-source library to perform the decoding and augmenting of images or speech signals. It provides the outer pipeline for the developers to process the raw data samples on the GPU devices on the parallel units. In practical scenarios, this GPU-friendly dataloader can effectively reduce the time consumption of training, and stabilize the communication frequency and interval between the CPUs and GPUs. It can be integrated into python, and many widely used deep learning frameworks, e.g. MXNet, PyTorch, and TensorFlow.
Continual learning is another fundamental research in deep learning. Different from the scenario of primitives for offline training, it allows the training process to be built on dynamic model architectures and unfixed streams of datasets. This training mode will cause large congestion under the most of current dataloaders. To tackle these difficulties, \citet{carta2023avalanche} provide and maintain the \textit{Avalanche}, an open source library under the PyTorch base, to expand the scenario of continual learning and incremental training. \textit{AvalancheDatasets} can be subsampled and concatenated to define and manipulate streams and replay buffers in the local training. It also provides a flexible setup of controlling the balance and cross-sampling from multiple datasets.

\subsection{System-centric Model Efficiency}
Hardware design and algorithm design are complementary processes. As for the researchers, based on their understanding of the model, they can optimize the algorithms for computation and memory based on the memory, computation, and communication features of the device, such as CPU, GPU, and other hardware architectures. In this part, we focus on the efficient storage of the deployment of large-scale models and reducing their practical training consumption with model parallelism.\\\\
\textit{Channel Last} is an efficient storage format implemented as an alternative to 4D \textit{NCHW} tensors in the training. This approach stores tensors pixel by pixel, leaving the channel as the last dimension. Under mixed precision training, models trained in channel-last format on a tensor core can increase training throughput~\citep{channelslast}. Another efficient method is adopting quantized low-precision to store and compute the tensors. \textit{Mixed precision training}~\citep{micikevicius2017mixed} is a technique that works in both single and multi-GPU situations which accelerates the network training process by using a mixture of single-precision and half-precision floating-point formats. When using FP16, i.e., half-precision floating-point numbers to store weights and gradients, the training is accelerated by increasing throughput while reducing memory usage and access. At the same time, an FP32 master copy of weights that updates with the weight gradient during the optimizer step is stored to maintain the network accuracy achieved by single-precision training. The use of mixed precision allows training larger models or batch sizes on a given hardware.
BF16~\citep{WangKanwar2020} is a different half-precision format than FP16 which can represent a larger range of floating point numbers to improve stability, and using it instead of FP16 may give better results~\citep{laurenccon2022bigscience, chowdhery2022palm}. Sparse training is also a good solution for the efficient training.
There is a lot of work on sparsity in models, but in real-world scenarios, sparse computation requires hardware support, otherwise, acceleration is difficult to achieve. Fortunately, advanced hardware designs have introduced more support for sparsity, such as NVIDIA's Ampere GPU architecture that introduces support for sparsity in its matrix math unit, i.e., the tensor core, which utilizes a 2:4 (50\%) sparsity pattern~\citep{mishra2021accelerating}, enabling twice the mathematical throughput of dense math. 
Structured pruning-based approaches leverage sparsity in hardware in CPU and GPU architecture, and can benefit the up-scaling of the models~\citep{jaszczur2021sparse}.
To achieve structured sparsity to fit the hardware, \citet{ma2023hrbp} propose Hardware-friendly Regrouping towards Block-based Pruning (HRBP), which preserves the extracted dense blocks in backpropagation.

For extremely large-scale models, model parallelism is an expected method which can efficiently reduce the training time by increasing the computational resources. For large models that cannot be put down on a single GPU card but can be put down on a multi-GPU server, or for an operation in a model that consumes a lot of GPU memory (e.g., softmax with a large number of classifications), tensor parallelism (also called intra-layer parallelism or model parallelism) can be used to reduce memory usage and average the computation. In tensor parallelism, some specific weights, gradients, and optimizer states of the model are split between devices, with each slice of the original tensor on its designated GPU, and computed in a distributed setup. 
Take the example of the MLP block of a transformer layer given in \cite{narayanan2021efficient}, which consists of two GEMMs and a GELU function, $Y=GeLU(XA), Z=YB$. Given two processors, we divide $A$ into $[A_1 A_2]$ by columns so that the output can be fed into $GeLU$ independently. $Y_i=GeLU(X A_i)$ is computed on each processor, resulting in $[Y_1\ Y_2]=[GeLU(X A_1)\ GeLU(X A_2)]$ . This is known as the column-parallel approach. When the second linear layer $Z=YB$ follows the column-parallel layer, we divide $B$ into $[B_1 B_2]^\top$ by rows to match the shape of $Y$ and avoid additional communication, which is called the row-parallel. To compute $Z=[Y_1 Y_2][B_1 B_2]^\top$, we first compute $Y_iB_i$ and then aggregate the result to $Z=Y_1B_1+Y_2B_2$ using the all-reduce method. 
We note that the column-parallel linear layer requires the application of an all-reduce across processors in the backward pass to aggregate the split gradient of the input tensor. For example, for $Y=XA$, all-reduce is applied to obtain $\dot X=\dot Y A^\top=\dot Y_1 A_1^\top+\dot Y_2 A_2^\top$.
However, when the model is extremely large that it must be split across multiple multi-GPU servers, there are two problems of the parallelism~\citep{narayanan2021efficient,smith2022using}: 
(a) tensor parallelism requires high-bandwidth communication between nodes such as NVLink within a single DGX A100 server, but for communication between multiple servers, the inter-server links are slower;
(b) more nodes lead to smaller matrix multiplications (GEMMs) which may reduce GPU utilization.

The hardware support from the model-centric perspective is mainly aimed at storage and calculations. Efficient storage allows the model to save a lot of time and achieve acceleration during transferring and allocating of devices. And, using mixed precision data types also can effectively avoid the redundancy of waiting time while maintaining the effect. Some of the underlying approaches based on module-specific or layer-specific optimization also contribute significant benefits, which provide a solid foundation for efficient training on large-scale models.

\subsection{System-centric Optimization Efficiency}
The optimization corresponds to the backpropagation and parameter updating process, which is the largest time-consuming calculation in the training. It is critical for the accelerations. The traditional paradigm has relatively limited operational space in the optimization process, influenced by the backpropagation mechanism. Therefore, one of the main determinants of training efficiency is the ability to achieve an efficient underlying optimization process. In this part, we mainly review the practical implementation including the rematerialization through computational graph rebuilding, pipeline parallelism, fused operators, efficient communication across devices, and Zero Redundancy Optimization~(ZeRO).\\\\
Rematerialization is a technique of rebuilding the entity in the computational graph which could efficiently avoid specific redundancy. It usually changes the gradient calculations.
Gradient checkpointing~\citep{chen2016training}~(rematerialization) can be used to re-build the computational graph, which releases the intermediate outputs and recomputes the forward results during the backpropagation to recover the released tensor. 
Rematerialization also could be implemented manually, as shown by \citet{touvron2023llama} in the efficient implementation of LLaMA. To improve the training efficiency, they store the output of linear layers by re-implementing the backpropagation via rebuilding the graph instead of using autograd. Their process reduces the costs of activations which are computationally expensive to recompute during the backpropagation process. 

Pipeline parallelism is an efficient parallel technique that decomposes incoming batches into mini-batches (which reduce the pipeline bubble), and divides the layers of the model across multiple GPUs, thus creating a pipeline where each stage, i.e. a set of contiguous layers takes the results of the previous stage as input and passes the results downstream, allowing different GPUs to participate in the computational process in parallel. It has the lowest communications and can be scheduled across nodes. 
State-of-the-art system~\citep{smith2022using} use a combination of reduced communication and increased locality and parallelism to improve training efficiency. \citet{huang2019gpipe} propose the Gpipe schedule in which the backward passes would be executed only when the full forward passes of a batch finish (F-then-B). \citet{narayanan2019pipedream} propose a scheduling mechanism one-forward-one-backward (1F1B), and \citet{narayanan2021efficient} show the 1F1B schedule with interleaved stages, where each device is assigned multiple subsets of layers (called model blocks) rather than a set of consecutive layers, which is more efficient in both time and space than F-then-B.
The limited memory of a single GPU limits the maximum model size it can accommodate, as well as the GPU utilization. To further increase the model size to get better accuracy, existing open source frameworks usually provide model parallel libraries to help efficiently distribute and train models on multiple nodes to improve accuracy and throughput for optimal distributed training. These frameworks integrate parallel training and collective communication libraries (such as NCCL), allowing researchers to address training bottlenecks by tuning within an existing training setup.

Fused implementations adopt the high efficiency of reusing the computational units. 
These fundamental optimizations can reduce memory I/O and improve compute efficiency by combining operations. Kernel fusion, for example, refers to combining multiple operations into a single kernel to improve performance and numerical stability.
By using deep compilers such as nvFuser for NVIDIA GPUs, researchers can fuse multiple neural network layers into a single CUDA kernel and can reconstruct CUDA operators to efficiently utilize the GPU's computational units and further reduce memory I/O. These approaches contribute significantly to the overall speedup. 
For example, NVIDIA Apex~\citep{apex} implements the fused versions of some common optimizers, such as Adam, which avoids multiple passes to and from GPU memory and is faster than the implementation in PyTorch.
PyTorch's recent 2.0 release provides \textit{torch.compile}, which can use the OpenAI Triton compiler to generate code that performs as well as custom hand-written kernels and dedicated CUDA libraries. 

Efficient communication is extremely important in training accelerations, especially in large-scale training. Communication bus efficiency across multi-nodes becomes the main bottleneck. Even in a single node, extensive communications between the CPU and GPU still lead to a serious obstruction.
In large-scale distributed deep learning training, it is important to focus on the efficiency of communication between GPUs, which communicate via PCIe or NVLink on a single machine and via Socket (Ethernet) or InfiniBand (with RDMA) between multiple machines, with higher communication bandwidth intra-node than inter-node.
Existing algorithms and systems can be optimized to improve communication efficiency.
For example,  \textit{1-bit compression}~\citep{seide20141} reduce the communication by quantizing gradients to as low as 1 bit per value. \citet{tang20211} propose \textit{1-bit Adam} which works with the non-linear gradient-based optimizer \textit{Adam}.
\citet{lu2022maximizing} propose \textit{0/1 Adam} to address the non-linearity challenges in \textit{Adam} when applying aggressive 1-bit quantization and local steps, achieving higher training throughput and end-to-end training time reduction compared to the \textit{1-bit Adam} baseline.
\citet{markov2021project} propose a communication framework named CGX, which supports compressed communication for data-parallel training.
In addition, better parallel strategies for efficient communication can be explored.
Data parallelism, tensor parallelism, and pipeline parallelism are the three commonly used parallel strategies. 
For example, \citet{narayanan2021efficient} show how to compose data, tensor, and pipeline parallelism to scale to thousands of GPUs. 

\citet{rajbhandari2020zero} propose the Zero Redundancy Optimizer (ZeRO), which is currently important technique for training large-scale models. ZeRO optimizes redundant model states (i.e. optimizer states, gradients, and parameters) in memory by partitioning them in three corresponding stages across processors and optimizing the communication, with the final model states evenly split on each node. 
On the second stage (ZeRO-2) which partitions both the optimizer states and the gradients, ZeRO-Offload \citep{ren2021zero} is built, which offloads the data and computations of both states to the CPU, thus leveraging the CPU to save the GPU memory. 
However, due to limited GPU memory, the number of parameters stored in GPU memory and copied over all devices is still limited, and sub-optimal data partitioning and limited PCIe bandwidth require large batch training for efficiency. 
Further, \citet{rajbhandari2021zero} show ZeRO-Infinity, a heterogeneous system technology that leverages CPU and NVMe memory (which is cheap, slow, but massive) in parallel across multiple devices to aggregate efficient bandwidth for current GPU clusters.
The current large-scale deep learning models often reach 10B or even 100B parameters, such as GPT3~\citep{brown2020language} (175 B), MT-NLG~\citep{smith2022using} (530 B). Training such large models usually require a comprehensive use of data parallelism, tensor parallelism, pipeline parallelism, mixed precision training, and ZeRO-like distributed optimizers, also known as 3D hybrid parallelism. 
Usually, tensor parallelism with the greatest communication volume is prioritized within a single node. 
Data parallelism is also placed within a node if possible to speed up the gradient communication, or mapped to a nearby node.

\subsection{Open Source Frameworks}

For researchers who want to apply parallel techniques to train large models simply, various existing open-source frameworks provide convenient implementations. 
To scale from single-GPU to multi-GPU or even multi-machine, it is often a simple matter of importing libraries, defining a training strategy and environment, and wrapping the optimizer. 
Below we introduce some other platforms, open source frameworks and libraries, and how they benefit (distributed) training of large-scale models.\\\\
JAX~\citep{jax2018github} is a library for machine learning which combines Autograd (an automatic differentiation library) and XLA (Accelerated Linear Algebra, a compiler for linear algebra), and provides the same API as NumPy that is compatible with GPU and TPU. 
For deep learning, especially in the case of using TPUs, JAX-based libraries such as Flax~\citep{flax2020github}, and Haiku~\citep{haiku2020github} provide deep learning features, and the efficient computation of hessian by JAX is also worth considering. There is a library Mesh-Transformer-JAX~\citep{mesh-transformer-jax} for parallel transformers in JAX and Haiku. In addition, Google recently released ViT-22B using JAX-based library Scenic~\citep{dehghani2021scenic}.
DeepSpeed~\citep{rasley2020deepspeed} is an open-source PyTorch library for large model training, which provides Zero Redundancy Optimizer (ZeRO)~\citep{rajbhandari2020zero,ren2021zero,rajbhandari2021zero}, 3D-parallelism, etc. and is used in BLOOM (176B)~\citep{scao2022language} and MT-NLG (530B)~\citep{smith2022using} training. 
DeepSpeed's Data Efficiency library~\citep{li2022deepspeed} combines curriculum learning and a random layerwise token dropping technique (random-LTD)~\citep{yao2022random}.
Recently, Microsoft launched the TorchScale open source toolkit~\citep{torchscale} for scaling up transformers.
Meta Research‘s FairScale~\citep{FairScale2021} is a PyTorch extension library that provides distributed training techniques, such as model parallelism (including the FullyShardedDataParallel, FSDP), activation checkpointing, gradient accumulation, gradient cropping, etc., to improve memory usage and speed up the training process. FairScale also includes some optimizers and schedulers, such as Adascale (large batch optimizer), FusedAdam, FusedSGD, etc., to improve the stability and efficiency of training large models. 
Another library Fairseq~\citep{ott2019fairseq} provides a series of sequence-to-sequence and language model implementations, such as Transformer, RoBERTa~\citep{liu2019roberta}, etc., and also provides some tools for data pre-processing and training.
There is also xFormers~\citep{xFormers2022}, a library of Transformer building blocks that supports the composable construction of Transformers.
The PaddlePaddle platform~\citep{ma2019paddlepaddle}is a framework designed for parallel distributed deep learning, supporting various deep learning hardware. Related development Kits and the series of work followed by ERNIE~\citep{sun2019ernie} cover a lot of training techniques.
\begin{table}[t]
\small
\centering
\caption{Frameworks for efficient training.}
\label{tab:frameworks}
\vskip 0.075cm
\setlength{\tabcolsep}{1.45mm}{\begin{tabular}{lll}
\toprule
Hosters & Libraries & Benefits \\
\midrule
PyTorch & PyTorch 2.0 & \textit{torch.compile}, mixed precision, DDP, and FSDP\\


Baidu & PaddlePaddle & 3D parallelism, mixed precision, GroupSharded (ZeRO-like), model implementations\\

ByteDance & LightSeq & mixed precision, fused operators, GEMM optimization \\

Google & JAX & Autograd, XLA\\

MicroSoft  & DeepSpeed & 3D parallelism, ZeRO, mixed precision, data efficiency library\\
                    & TorchScale & scaling up Transformers \\

MetaResearch & FairScale & FSDP, pipeline parallel, general techniques, Adascale \\
           & Fairseq    & FSDP, general techniques, model implementations\\
           & xFormers & Transformer model implementations\\

MosaicML & Composer & multi-GPUs training, implementations of speedup methods\\
           
Huggingface & Transformers & model implementations \\
                        & Accelerate & distributed training, mixed precision, support of FSDP, DeepSpeed, and Megatron-LM\\

HPC-AI Tech & Colossal-AI & 3D(especially for tensor) parallelism, sequence parallelism, ZeRO\\

OneFlow & LiBai & 3D parallelism, mixed precision, ZeRO, general techniques\\

NVIDIA & Megatron-LM & distributed pretraining, model implementations \\
& Apex & AMP, distributed training, fused optimizers, fused layers, and layer norm \\

\bottomrule
\end{tabular}}
\end{table}\\
Some provide comprehensive operations on the entire process.
Megatron-LM is the large transformer lib developed by NVIDIA, with a efficient training framework. Also, NVIDIA maintains the Apex~\citep{apex} library of tools for mixed precision and distributed training.
MosaicML's Composer~\citep{mosaicml2022composer}, a PyTorch library for efficient training which creates an interface for each part of the training process, allowing composing various speedup methods both built-in or customized into the training pipeline to develop efficient training recipes. 
ByteDance LightSeq~\citep{wang2021lightseq,wang2021lightseq2}, an efficient inference and training engine for Transformers.
Bagua~\citep{gan2021bagua} developed by Kuaishou and ETH Zürich, a deep learning training acceleration framework for PyTorch.
Colossal-AI~\citep{bian2021colossal}, which provides a series of parallel components to build and train distributed AI models. In addition to the 1D tensor parallelism mentioned above, the Colossal-AI~\citep{bian2021colossal} platform also integrates 2D, 2.5D, 3D tensor parallelism strategies \citep{xu2021efficient,wang20212,bian2021maximizing} and sequence parallelism~\citep{li2021sequence} that distribute computational and memory load and optimize communication costs further. 
%
EleutherAI GPT-NeoX~\citep{gpt-neox-library}, a library for training large-scale language models on which a 20 billion parameter model GPT-NeoX-20B~\citep{gpt-neox-20b} is built.
Ray~\citep{ray_project}, an open-source framework for scaling AI and Python applications like machine learning.
The fastai deep learning library built by \citet{howard2020fastai}, which provides researchers with components both at high and low levels to build state-of-the-art or configurable approaches.
OpenBMB BMTrain~\citep{bmtrain}, a large model training toolkit for efficient training of big models.
%
OneFlow-LiBai~\citep{of2021libai}, which is a large-scale open-source training toolbox that provides 3D-parallelism distributed training and other training techniques.
Huawei MindSpore~\citep{mindspore}, an AI computing framework. 
%
Horovod~\citep{sergeev2018horovod} hosted by LF AI \& Data Foundation, a distributed deep learning framework.

In summary, we organize Table~\ref{tab:frameworks} to summarize the benefits that each framework\ library and clearly show their main contributions to efficient training. As we have mentioned, training large models requires a comprehensive use of general techniques. Several deep learning frameworks have been designed to provide efficient support for training accelerations. We focus on valid implementation which could help to achieve higher efficiency. More techniques are still emerging and being explored with prolonged developments.

\subsection{Discussion}
In this section, we review the training accelerations from the system-centric perspective. The system implementation carries the algorithm design which determines the training efficiency in the practical scenarios. The efficiency of the designed algorithms and techniques also relies heavily on sound hardware support for maximum performance. In this section, we study the current techniques of data processing, model training, and optimization execution at the system level. We summarize the advantages and disadvantages of existing methods as follows.

In terms of system-centric data processing, the key routes on this include high-performance data processing and data parallelism. As the volume of the dataset grows rapidly, the pre-processing of the training data samples becomes complicated and onerous. This is very detrimental to the development of large batch training. When data processing time does not match training consumption, the training process goes into a waiting state, resulting in very low efficiency. More and more research tends to reduce communication and maintain the localization of data processing, yielding high-efficient processing and parallelism. The same idea has been applied at the model level. Different from data processing, models are usually stored on computing devices. Therefore, we have to face the challenge of the deployment of large-scale models. A range of quantitative compression-based techniques and mixed precision models effectively guarantee the training accelerations. In addition, in large-scale training, the optimization of distributed architecture also helps to improve the computational efficiency of the models. In addition to data processing and model deployment, hardware support for the optimization process is the most important part of practical training. Current mainstream technology routes revolve around the efficient calculations on the reconstructed computational graph, pipeline parallelism for stage-wise training, fused implementation of the efficient optimizers, and efficient communication across the multi-nodes and multi-devices. Graph and pipeline determine the simplified calculation flow on the hardware calculations, and communication efficiency resolves the systemic wall-clock time redundancy in the training. With the sound support of the hardware system, the effect of the designed algorithms could be performed to the fullest extent. 

In the future, we see the light in the following research directions:
\begin{itemize}
    \item Localized data processing. We can make full use of the efficient parallel computation in the GPU devices to achieve pre-processing such as data regularization in large batch training. There are some efficient libraries to achieve this technique, but it has not been fully popularized yet. A lot of operations are still to be discovered.
    \item Vertical Model Deployment. As the model size grows larger, the limitations of computing devices will become more apparent. Meanwhile, the calculation of individual modules in the large-scale model cannot be fully parallelized due to the chain rules of the gradients. Research on vertical training on multi-nodes multi-devices becomes a promising direction worth exploring. The optimization algorithms of vertical training are also expected further studies.
    \item Novel efficient paradigm. Developing novel and efficient paradigms can further accelerate training for large models. It should jointly consider both the acceleration techniques of each individual module and their potential interactions and conflicts, which could provide a vigorous implementation of the designed framework for efficient training. For instance, considering the large-scale edged computational devices, collaboratively training models in decentralized setups is also a promising study in the future~\citep{yuan2022decentralized}.
\end{itemize}

 \section{Conclusion}\label{sec:conclusion}

In this survey, we review the general training acceleration techniques for efficiently training large-scale deep learning models. We consider all the components in the \textit{gradient-based} update formula~(\ref{eq:gradient-based descent}) which cover the total training process in the field of deep learning. We present a novel taxonomy to summarize and categorize these techniques into five major directions: ``Data-centric", ``Model-centric", ``Optimization-centric", ``Budgeted training", and ``System-centric". In the first four parts, we mainly focus on comprehensive studies from the perspectives of algorithm design and methodology. In the “system-centric efficient training” part, we summarize the practical implementation from the perspectives of paradigm innovation and hardware support.  We review and summarize the commonly used or latest developed techniques corresponding to each component, their benefits and trade-offs of proposed techniques, and discuss their limitations and promising future studies for efficient training.

While providing a comprehensive technical review and guidance, our survey also presents current breakthroughs and bottlenecks in effective training. We hope that this survey will not only help researchers efficiently achieve general training accelerations, but also provide some meaningful and promising implications for the future development of efficient training. 
In addition to some of the potential advances mentioned at the end of each section, here we put forward some broader and promising perspectives: 
\textbf{(i) Efficient Profile Search.} Efficient training could design the prebuilt and customizable profile search strategies for the models from the perspectives of data augmentation combinations, model architectures, optimizer designs, etc. There has been some progress in this study. Furthermore, it also deserves exploration as to new model structures and compression patterns, new pre-training tasks, and the exploitation of ``model-edge"~\citep{han2021pre} knowledge. In the future, it is sure to achieve more brilliant successes. 
\textbf{(ii) Adaptive Scheduler}. Using an optimized scheduler, such as for curriculum learning, learning rate and batch size, and model complexity, it is possible to achieve better performance. 
Budget-aware schedulers that dynamically adapt to the remaining budget reduce the cost of manual design in these aspects. 
Adaptive schedulers could be used to explore more powerful parallelism and communication methods, taking into account more general and practical scenarios, e.g. on the heterogeneous networks across wide regions and data centers for large-scale decentralized training. 


\bibliographystyle{plainnat} 
\bibliography{reference.bib}

\end{document}